%% file: main.tex
\documentclass[journal]{IEEEtran}

\let\bibhang\relax
\usepackage[numbers,compress]{natbib}

\usepackage{setspace}
\usepackage{enumitem}
\usepackage{subfig}
\usepackage{booktabs}
\usepackage{xcolor}
\usepackage{tabularx}
\usepackage{graphicx}%
\usepackage{amsmath}
\usepackage{amssymb}
\usepackage{listings}
\usepackage{textcomp}
\usepackage{url}
\usepackage{multirow}
\usepackage{hyperref}
\usepackage{cleveref}
\usepackage{wrapfig}
\usepackage{mathtools}
\crefformat{footnote}{#2\footnotemark[#1]#3}
\usepackage{ragged2e} 
\usepackage[subnum]{cases}
\newcolumntype{Y}{>{\RaggedRight\arraybackslash}X} 
\usepackage{algorithm}
\usepackage{bm}
\usepackage[noend]{algpseudocode}
\usepackage{hyperref}
\usepackage[utf8]{inputenc}
\usepackage{tcolorbox}
\usepackage[font=small]{caption}
\usepackage{balance}

\newcommand{\eg}{\emph{e.g.},}
\newcommand{\ie}{\emph{i.e.},}
\newcommand{\etal}{\emph{et~al.}}

\DeclarePairedDelimiterX{\norm}[1]{\lVert}{\rVert}{#1}
\def\secref#1{Sec.\ \ref{#1}}
\def\figref#1{Fig.\ \ref{#1}}
\def\tabref#1{Table~\ref{#1}}

\graphicspath{ {figures/Experiments_and_Results/} {figures/Hybrid_Control_Architecture/}}

\newcommand{\degree}{$^{\circ}$}

\newcommand{\flatpack}{\textit{FlatPack}}
\newcommand{\spinningpack}{\textit{SpinningPack}}

\usepackage{pifont}%
\newcommand{\cmark}{\ding{51}}%
\newcommand{\xmark}{\ding{55}}%

\newtcbox{\inlinecode}{on line, boxrule=0pt, boxsep=0pt, top=2pt, left=2pt, bottom=2pt, right=2pt, colback=white, colframe=white, fontupper={\ttfamily \small}}

\usepackage{todonotes}

\input{macros}

\begin{document}

\title{Wildcat: Online Continuous-Time \\3D Lidar-Inertial SLAM}

\author{
Milad Ramezani,~\IEEEmembership{Member,~IEEE,}, Kasra Khosoussi, Gavin Catt,~\IEEEmembership{Member,~IEEE,} Peyman Moghadam,~\IEEEmembership{Senior Member,~IEEE,} Jason Williams,~\IEEEmembership{Senior Member,~IEEE,} Paulo Borges,~\IEEEmembership{Senior Member,~IEEE,}   Fred Pauling,~\IEEEmembership{Member,~IEEE},  Navinda Kottege,~\IEEEmembership{Senior Member,~IEEE}

\IEEEcompsocitemizethanks{\IEEEcompsocthanksitem $^1$ The authors are with the Robotics and Autonomous Systems Group, DATA61, CSIRO, Brisbane, QLD 4069, Australia.
E-mails: {\tt\small \emph{firstname.lastname}@csiro.au}}
}

\markboth{IEEE Transactions on Robotics ,~Vol.~xx, No.~x, xx~2022}%
{Shell \MakeLowercase{\textit{et al.}}: Bare Demo of IEEEtran.cls for IEEE Journals}

\maketitle

\begin{abstract}
\input{chapters/abstract.tex}

\end{abstract}

\begin{IEEEkeywords}
3D Lidar-Inertial SLAM, Localisation and Mapping, Collaborative SLAM
\end{IEEEkeywords}

\input{chapters/introduction.tex}

\input{chapters/related_work.tex}

\input{chapters/overview.tex}

\input{chapters/odom.tex}

\input{chapters/pgo.tex}

\input{chapters/experiments.tex}

\input{chapters/conclusion.tex}

\section*{Acknowledgement}
The authors would like to thank the engineering team in the Robotics and Autonomous Systems Group, CSIRO Data61 for their support. We would also like to thank AutoMap\footnote{https://automap.io/} for their help in target selection using AutoMap software. Technology described herein is the subject of International PCT Patent Application No.: PCT/AU2021/050871 entitled ``Multi-Agent Map Generation", filed in the name of Commonwealth Scientific and Industrial Research Organisation on August 09, 2021.

\balance

\bibliographystyle{IEEEtranN}
\bibliography{references}

\end{document}

%% file: macros.tex
\newcommand{\Wcal}{\mathcal{W}}
\newcommand{\Mcal}{\mathcal{M}}
\newcommand{\Rset}{\mathbb{R}}
\newcommand{\xx}{\mathbf{T}}
\newcommand{\rr}{\mathbf{r}}
\newcommand{\rc}{\rr^\text{cor}}
\newcommand{\RR}{\mathbf{R}}
\newcommand{\Rc}{\mathbf{R}^\text{cor}}
\newcommand{\pp}{\mathbf{t}}
\newcommand{\pc}{\pp^\text{cor}}
\newcommand{\xc}{\xx^\text{cor}}

\newcommand{\nn}{\mathbf{n}}
\newcommand{\grav}{\mathbf{g}}
\newcommand{\su}{\mathbf{p}}
\newcommand{\bb}{\mathbf{b}}
\newcommand{\ac}{\mathbf{a}}
\newcommand{\om}{\boldsymbol\omega}
\newcommand{\linterp}{\mathsf{LinInterpolate}}
\newcommand{\rotlinterp}{\mathsf{RotInterpolate}}
\newcommand{\Timu}{\mathcal{T}^\text{imu}}

%% file: chapters/abstract.tex
We present Wildcat, a novel online 3D lidar-inertial SLAM system with exceptional versatility and robustness. At its core, Wildcat combines a robust real-time lidar-inertial odometry module, utilising a continuous-time trajectory representation, 
with an efficient pose-graph optimisation module that seamlessly supports both the single- and multi-agent settings.
The robustness of Wildcat was recently demonstrated in the DARPA Subterranean Challenge where it outperformed other SLAM systems across various types of sensing-degraded and perceptually challenging environments.
In this paper, we extensively evaluate Wildcat in a diverse set of new and publicly available real-world datasets and showcase its superior robustness and versatility over two existing state-of-the-art lidar-inertial SLAM systems.

%% file: chapters/introduction.tex
\section{Introduction}
\label{sec:intro}

\begin{figure}[tp]
\centering
\subfloat{\includegraphics[width=1\linewidth]{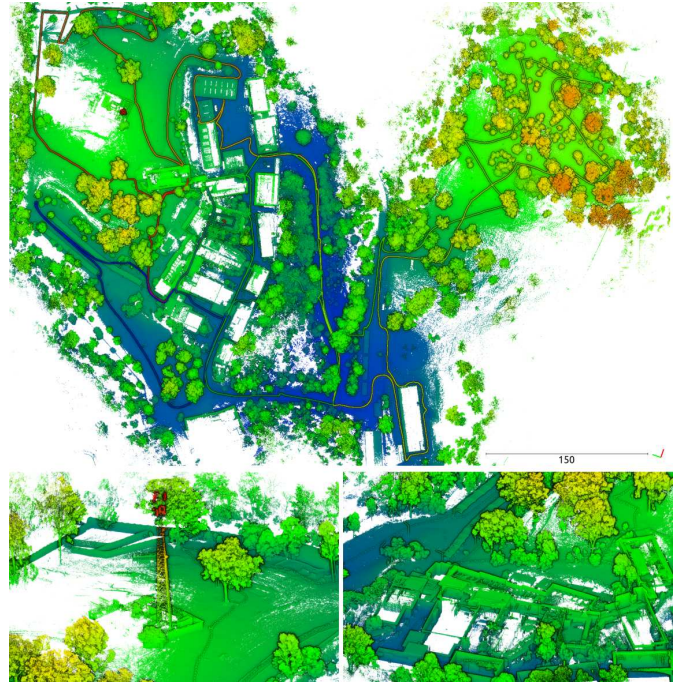}}

\caption{\small{\textbf{Top}: Wildcat can map large-scale environments, including various types of structured and unstructured areas indoors and outdoors. \textbf{Bottom}: Two close-up views from the challenging areas where the state-of-the-art systems struggle to converge (see~\secref{sec:map}). 
}}
\label{fig:frontpage}
\end{figure}

\IEEEPARstart{S}{imultaneous} Localisation and Mapping (SLAM) is the backbone of robotics downstream tasks such as
robot navigation in unknown GPS-denied environments.  %
Among existing solutions to SLAM,
lidar-inertial systems are highly popular due to their robustness, precision,
and high-fidelity maps. Beyond robotics applications, these systems also hold
the promise of providing scalable and low-cost alternative to conventional mapping and surveying
systems (used in~\eg~construction) with comparable precision. 
During the last two decades and with the advent of affordable 3D lidars,
many 3D lidar-inertial SLAM systems have been proposed; see \cite{xu2021fast,BosZlo12,shan2020lio,behley2018efficient,park2018elastic} and references
therein. 

Despite tremendous progress in recent years, designing a
robust and versatile lidar-inertial SLAM system  remains a challenge.
In particular, designing features that can be reliably detected and matched in
a wide range of environments  
is a difficult task.
Additionally, lidar SLAM systems must be able to account for the effects of platform's motion and
the mechanical actuation within the sensor on lidar points (\ie{} motion distortion).
Although this issue can be mitigated by incorporating data from an Inertial Measurement
Unit (IMU), fusion of asynchronous data from lidar and IMU presents additional technical
challenges.

This paper presents Wildcat, a state-of-the-art online 3D lidar-inertial
SLAM system, and showcases its exceptional versatility and robustness over prior
state-of-the-art systems through extensive experimental evaluation and carefully
designed case studies. At its core, Wildcat combines an \emph{online}
implementation of concepts from the pioneering (albeit offline) odometry system proposed in
\cite{BosZlo12} with a pose-graph optimisation module, efficiently allowing to map large-scale environments as seen in~\figref{fig:frontpage}.
Thanks to its modular design, Wildcat also seamlessly supports decentralised
collaborative multi-agent localisation and mapping where agents exchange their submaps via peer-to-peer communication and independently optimise the
collective pose graph.

Wildcat has been heavily field tested (i) in various types
of environments such as buildings, urban roads, mines,
caves, farms and forests; and (ii) on various types of platforms including handheld, ground
vehicles (\eg~cars, legged robots, and tracked vehicles), and aerial robots.
Most recently, the robustness and versatility of Wildcat were demonstrated in the
DARPA Subterranean Challenge where it outperformed other state-of-the-art SLAM
systems in a wide range of perceptually challenging and sensing-degraded 
subterranean environments (\eg~due to dust and smoke). Specifically, it was
reported by DARPA that the map produced by Wildcat using a team of four heterogeneous robots in the Final
Event had
``0\% deviation'' and ``91\% coverage'',\footnote{\url{https://youtu.be/SyjeIGCHnrU?t=1932}} where deviation
is defined as the percentage of points in the submitted
point cloud that are farther than one meter from the points in
the surveyed point cloud map.

The main contributions of this paper are the following:
\begin{itemize}
		\item We present Wildcat, a highly robust and versatile state-of-the-art lidar-inertial SLAM system. This paper provides a detailed technical description of Wildcat beyond the broad non-technical overview previously presented in \cite{hudson2021heterogeneous}.

		\item We demonstrate the robustness and versatility of Wildcat through
				carefully designed experiments. This includes quantitatively
				comparisons against two 
				other state-of-the-art lidar-inertial systems
				\cite{shan2020lio,xu2021fast2} on a publicly available dataset
				\cite{Kim2020MulRanMR} and two unique new large-scale
				multi-domain datasets with over 60 accurately
				surveyed landmarks.
\end{itemize}

\subsection*{Outline}
The remainder of this paper is organised as follows. We first introduce the
necessary notation below. We then review existing lidar-based SLAM systems in
Section~\ref{sec:related-work}. An overview of Wildcat's core components is presented in
Section~\ref{sec:overview}. This is followed by a detailed description of
Wildcat's odometry and pose-graph optimisation modules in
Sections~\ref{sec:odom} and \ref{sec:pgo}, respectively. The results of extensive experimental
evaluations and quantitative comparisons against the state of the art are
presented in Section~\ref{sec:experiments}. Finally, we conclude the paper in
Section~\ref{sec:conclusion} where we discuss future research
directions.

\subsection*{Notation}
We use $[n] \triangleq \{1,2,\ldots,n\}$ to refer to the set of natural
numbers up to $n$.
Bold lower- and upper-case letters are generally reserved for
vectors and matrices, respectively.
The standard inner product on $\Rset^n$ is written as $\langle
\cdot,\cdot\rangle$.
The special Euclidean and special orthogonal groups are denoted by
$\mathrm{SE}(3)$ and $\mathrm{SO}(3)$, respectively. 
We use $\mathfrak{so}(3)$ to refer to the Lie algebra associated
to
$\mathrm{SO}(3)$.
Matrix exponential and logarithm are denoted by $\exp$ and
$\log$, respectively. The hat operator
$(\cdot)^\wedge : \Rset^3 \to \mathfrak{so}(3)$ gives the
natural representation of vectors in $\Rset^3$ as $3 \times 3$ skew-symmetric matrices.
The inverse of hat operator is denoted by $(\cdot)^\vee :
\mathfrak{so}(3) \to \Rset^3$.
Finally, the linear-interpolation operator
$\linterp$ is defined as follows,
\begin{align}
		\linterp & : \Rset^n \times \Rset^n
		\times [0,1] \to \Rset^n\nonumber\\
		(\mathbf{x},\mathbf{y},\alpha) & \mapsto
		\alpha \mathbf{x} + (1-\alpha)\mathbf{y}.
		\label{eq:linearInterp}
\end{align}

%% file: chapters/related_work.tex
\section{Related work}
\label{sec:related-work}

One of the most popular and influential lidar-inertial-based systems is Lidar Odometry and Mapping (LOAM) \cite{ZhaSin14,ZhaSin17}.  Assuming constant angular and linear velocity during a sweep, LOAM linearly interpolates the pose transform at a high frequency (10\,Hz) but with low fidelity over the course of a sweep. By minimising the distance between corresponding edge point and planar point features extracted in one sweep and the next sweep as evolving, ego-motion is estimated iteratively until convergence. Later, at a lower rate (1\,Hz), features of the frontier sweep which are deskewed by the odometry algorithm are matched with the map generated on the fly to estimate the sweep pose in the map frame. In \cite{wang2021f} the authors propose a computationally efficient framework based on LOAM that can run at 20\,Hz. It deals with lidar distortion in a non-iterative two-stage method by first computing the distortion from frame-to-frame matching and then updating it in the frame-to-map matching step once the current pose is estimated in an iterative pose optimisation. In contrast to the loosely coupled approach proposed in LOAM, LIO-Mapping~\cite{ye2019tightly} utilises a \emph{tightly-coupled} method in which lidar and inertial measurements are fused in a joint optimisation problem.

 LeGo-LOAM~\cite{shan2018lego} and LIO-SAM~\cite{shan2020lio} are two popular lidar-inertial SLAM systems based on LOAM that use pose-graph optimisation. Pose-graph optimisation enables these methods to remove drift due to accumulated odometry error and create globally consistent maps by incorporating loop closures. In particular, LIO-SAM is a tightly-coupled keyframe-based online system that combines lidar odometry with IMU preintegration, loop closures, and (if available) GPS factors via pose-graph optimisation. LT-Mapper~\cite{kim2021lt} builds on LIO-SAM and uses Scan Context (SC)~\cite{gkim-2018-iros} for loop closure detection. 
 We compare Wildcat with LIO-SAM in~\secref{sec:experiments}.
Another tightly-coupled lidar-inertial SLAM system is IN2LAMA~\cite{le2019in2lama}, an offline system that addresses the lidar distortion problem by interpolating IMU measurements using Gaussian Process (GP) regression.

Filtering-based frameworks such as LINS~\cite{qin2020lins}, FAST-LIO~\cite{xu2021fast}, and its successor FAST-LIO2~\cite{xu2021fast2} tightly couple IMU and lidar measurements. In particular, FAST-LIO2~\cite{xu2021fast2} is an odometry and mapping system based on iterated Kalman filter in which raw lidar points are efficiently registered to the map. FAST-LIO2 uses the backward propagation step introduced in \cite{xu2021fast} to mitigate lidar distortion.
We compare our method against FAST-LIO2 in~\secref{sec:experiments}.

Another line of work explores the idea of using complex and expressive continuous-time (CT) representations of robot trajectory to address the lidar distortion problem (by querying robot pose at any desired time), and also to facilitate fusion of asynchronous measurements obtained from sensors at different rates such as IMU and lidar~\cite{BosZlo12,alismail2014continuous,patron2015spline,furgale2013unified}; see \cite{park2021elasticity, park2018elastic} and references therein for a discussion of various CT-based frameworks. In particular, B-splines \cite{bosse2009continuous,BosZlo12,lovegrove2013spline} and approaches based on GP regression \cite{tong2013gpgn} are two popular choices. 
For example, \cite{tong2014pose} applies a GP-based approach~\cite{tong2013gpgn} to obtain the robot poses at measurement times in a lidar-based visual odometry system. 
Recently, \citet{droeschel2018efficient} propose a CT  hierarchical method for 3D lidar SLAM that, similar to our work, combines local mapping with pose-graph optimisation. The authors in \cite{droeschel2018efficient} use Spline Fusion~\cite{lovegrove2013spline} to address the lidar distortion problem. 
More recently, Park \etal{} \cite{park2021elasticity} proposed a map-centric SLAM system (ElasticLiDAR++) which uses a CT map deformation method to maintain a globally consistent map without relying on global trajectory optimisation. 
Our lidar-inertial odometry is an online  implementation of the concepts introduced in the offline systems proposed by Bosse and Zlot~\cite{bosse2009continuous} and Bosse~\etal~\cite{BosZlo12} based on cubic B-spline interpolation.

%% file: chapters/overview.tex
\section{Wildcat Overview}
\label{sec:overview}

\begin{figure*}[t]
		\centering
		\includegraphics[width=0.9\textwidth]{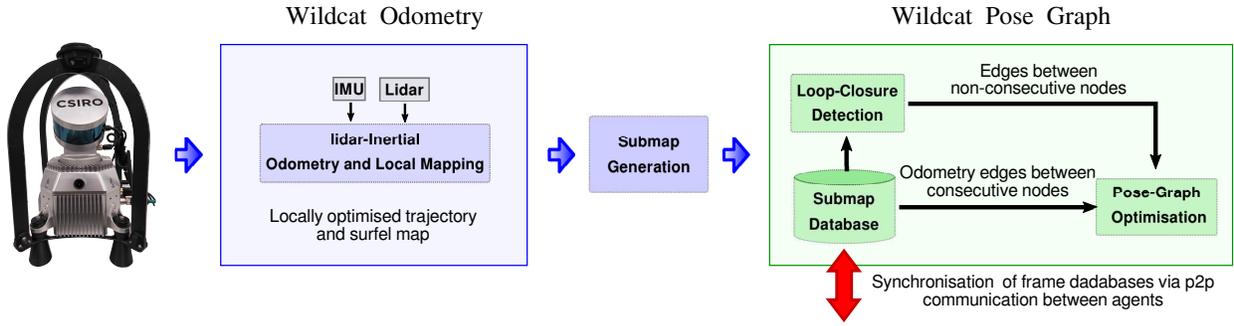}
		\caption{\small{Wildcat consists of two major modules: Wildcat odometry in which IMU and lidar measurements are integrated in a sliding-window fashion to continuously estimate the robot trajectory and produce submaps for pose-graph optimisation; Pose-graph optimisation where submaps are used to generate odometry edges as well as loop closures to efficiently map large-scale environments while correcting the odometry drift. Additionally, Wildcat can be used in multi-agent scenarios by sharing submaps via peer-to-peer (p2p) communication between the agents and optimising the collecting pose graph.} }
		\label{fig:flowchart}
\end{figure*}
Wildcat is composed of the following two main modules:
\begin{enumerate}
		\item A sliding-window lidar-inertial odometry and local
				mapping module (see Section~\ref{sec:odom}), hereafter referred to as \emph{Wildcat odometry}. Wildcat odometry is designed to efficiently integrate asynchronous IMU and lidar measurements using continuous-time representations of trajectory and to mitigate the distortion in map due to the motion of the sensor. 
		\item A modern pose-graph optimisation (PGO) module. By leveraging the odometry solution and local maps (Wildcat \emph{submaps}) produced by Wildcat odometry, the robot trajectory and environment map are optimised at a global scale. Wildcat merges submaps with sufficient overlap to reduce pose-graph nodes, effectively mapping large-scale environments
				(see Section~\ref{sec:pgo}).
				
\end{enumerate}
\figref{fig:flowchart} displays Wildcat's pipeline when running on a single
agent. In the following sections, we describe each module in
detail.

%% file: chapters/odom.tex
\section{Wildcat Odometry}
\label{sec:odom}

Wildcat odometry is a real-time implementation
of a number of concepts from \cite{BosZlo12}.
This module processes data in a sliding-window fashion. The $k$th time window
$\Wcal_k$ is a fixed-length time interval obtained by sliding the previous time
window $\Wcal_{k-1}$ forward by a fixed amount.  We now describe the key steps
taken by Wildcat odometry during the $k$th time window $\Wcal_k$. 

\subsection{Surfel Generation}
\label{sec:surfel}
We denote the true pose of robot at any arbitrary time $t$ by $\xx(t) = (\RR(t),\pp(t)) \in \mathrm{SO}(3)
\times \Rset^3$. 
Let  $\Timu_k \subset
\Wcal_k$ denote the set of timestamps of IMU measurements received within
$\Wcal_k$. After sliding forward the previous time window $\Wcal_{k-1}$, we initially estimate robot
poses at the timestamp of new IMU measurements~\ie~ $\{ \xx(t) :
t\in \Timu_{k} \cap (\Wcal_k \setminus \Wcal_{k-1})\}$, by
integrating new accelerometer and gyro measurements. %
We then perform linear
interpolation between these poses on $\mathfrak{so}(3) \times \Rset^3$ to
initialise 
robot poses associated to new lidar measurements (\ie~those received in
$\Wcal_k \setminus \Wcal_{k-1}$).
For each new lidar measurement,
interpolation is performed between the two closest (in time) IMU poses.
This initial guess is then
used to place raw lidar measurements in the world frame. 

Next, we generate surfels (surface elements) by clustering points based on their
positions and timestamps and fitting ellipsoids to them. First, we divide the
space into a set of cube-shaped cells (voxels) and cluster lidar points
within each cell and with proximal timestamps together. We then
fit an ellipsoid to each sufficiently large cluster of points based on a
predetermined threshold. The center of each ellipsoid (position of surfel) and its
shape are determined by the sample mean and covariance of 3D points in the
cluster, respectively.  
Recall
that lengths of an ellipsoid's principal semi-axes are given by the reciprocals
of the square root of the eigenvalues of the corresponding covariance matrix.
Therefore, the larger the gap between the two smallest eigenvalues of the
covariance matrix is (normalised by the trace), the ``more planar'' the
corresponding ellipsoid.  We thus quantify planarity of a surfel by computing a
score based on the spectrum of its covariance matrix
\cite[Eq.~4]{bosse2009continuous} and only keep surfels that are sufficiently
planar.  Further, we use the eigenvector corresponding to the smallest
eigenvalue of the covariance matrix to estimate the surface normal. Finally, we employ a
multi-resolution scheme where the clustering and surfel
extraction steps are repeated at multiple spatial resolutions (voxel sizes).

\subsection{Continuous-Time Trajectory Optimisation}
\begin{figure}[t]
\centering
\includegraphics[width=1.0\linewidth]{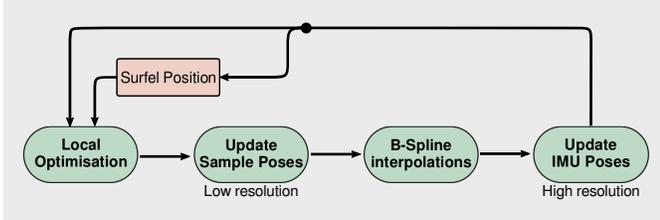}
\caption{\small{Continuous-Time optimisation framework in Wildcat. \emph{Local Optimisation} is based on multiple cost terms mainly  from IMU measurements and surfel correspondences. Local optimisation estimates a set of discrete poses (\emph{Update Sample Poses}), then in \emph{B-spline Interpolations}, by fitting a cubic B-spline over the corrected samples and the robot poses obtained from the IMU poses at the sample times, the robot poses (\emph{Update IMU Poses}) are updated at a high temporal resolution. This process is repeated iteratively to obtain an accurate estimate of robot trajectory and remove distortion from the surfel map.}}
\label{fig:odom_diagram}
\end{figure}

After generating surfels from \emph{new} lidar points in $\Wcal_k$, Wildcat odometry
performs a fixed number of \emph{outer} iterations by alternating between
(i) matching surfels given the current trajectory estimate,
and (ii) optimising the robot's trajectory using IMU measurements and matched
surfels and updating surfels' positions, which are described in Sections~\ref{sec:correspondence} and
\ref{sec:optimization}, respectively.~\figref{fig:odom_diagram} depicts the procedure of trajectory optimisation in Wildcat.

\subsubsection{Surfel Correspondence Update}
\label{sec:correspondence}
Each surfel is described by its estimated position in the world frame, normal
vector and resolution.
This information is used to
establish correspondences between surfels.  Specifically, we conduct
 k-nearest neighbour search 
 in the $7$-dimensional descriptor space for surfels created within the current time window $\Wcal_k$ and keep reciprocal matches whose average timestamps are farther than a small predefined threshold. We denote the set of matched surfel pairs by $\Mcal$. 

\subsubsection{Trajectory Update}~\\
\label{sec:optimization}
\noindent\textbf{Step 2.1 - Updating Sample Poses:} 
Let $\{t_i\}_{i=1}^n$ be the set of $n$ equidistant timestamps sampled from
$\Wcal_k$, and $\hat{\xx}(t_i) = (\hat{\RR}(t_i),\hat{\pp}(t_i))$ denote the estimate of robot
trajectory at the sampling time $t_i$. 
In this step, we first aim to compute small pose corrections
$\{\xc_i\}_{i=1}^{n} = \{(\Rc_i,\pc_i)\}_{i=1}^n \in (\mathrm{SO}(3) \times
\Rset^3)^n$ to update our estimate of robot's trajectory at sampling times
according to $(\hat{\RR}(t_i),\hat{\pp}(t_i)) \leftarrow (\Rc_i
\,\,\hat{\RR}(t_i), \pc_i + \hat{\pp}(t_i))$. We obtain these correction poses
by solving the following local optimisation problem,
\begin{equation}
		\underset{\substack{\xc_i \in \mathrm{SO}(3) \times \Rset^3\\
		i = 1,\ldots,n}}{\text{minimise}}
		\sum_{t \in \Timu_k} f_\text{imu}^t + \sum_{(s,s') \in \Mcal} 
		f_\text{match}^{s,s'}.
		\label{eq:optproblem}
\end{equation}
The cost functions $f_\text{imu}^t$ and
$f_\text{match}^{s,s'}$ (associated to IMU measurements and matched surfel
pairs, respectively) are functions of (small) subsets of
$\{\xc_i\}_{i=1}^{n}$ and their arguments are omitted for notation simplicity.
Before defining these cost functions in Section~\ref{sec:costs}, we describe key steps taken by Wildcat odometry to
update its estimate of robot's trajectory.
To solve \eqref{eq:optproblem}, we use a retraction to formulate the problem as an optimisation problem over 
$(\mathfrak{so}(3) \times \Rset^3)^n \cong (\Rset^6)^n$. This allows us to
express the constrained optimisation problem in \eqref{eq:optproblem} as an unconstrained
problem whose decision variables are $\{(\rc_i,\pc_i)\}_{i=1}^{n}
\in (\Rset^3 \times \Rset^3)^n$ such that $\Rc_i = \exp((\rc_i)^\wedge)$.
As we see shortly, this optimisation problem is a standard nonlinear least squares
problem which we solve (approximately) using Gauss-Newton.
Specifically, we  
linearise the residuals and
solve the normal equations to obtain $\{(\rc_i,\pc_i)\}_{i=1}^{n}$. We then
update our estimate of robot pose at sampling times according to 
$(\hat{\RR}(t_i),\hat{\pp}(t_i)) \leftarrow
(\exp((\rc_i)^\wedge) \,\hat{\RR}(t_i), \pc_i +
\hat{\pp}(t_i))$ for $i \in [n]$. We make this process robust to outliers using an
iteratively reweighted least squares (IRLS) scheme based on the Cauchy
M-estimator \cite{zhang1997parameter}.

\noindent\textbf{Step 2.2 - Updating IMU Poses:} 
To be able to solve the above optimisation problem sufficiently fast, the number of sample
poses $n$ is typically an order of magnitude smaller than the number of IMU
measurements in $\Wcal_k$.  However, Wildcat odometry must maintain an estimate of
robot's trajectory at a higher rate (say, 100 Hz) to accurately place surfels
in the world frame (see Step 2.3) and also for defining the cost functions
in \eqref{eq:optproblem} (see Section~\ref{sec:costs}).
We therefore use our
corrected sample poses to update our estimate of robot's trajectory at IMU
timestamps,~\ie~$\{\hat{\xx}(t) : t \in \Timu_k \}$.  Since the timestamps of
IMU measurements are not aligned with those of sample poses, we first use a cubic
B-spline interpolation between corrected sample poses
$\{\hat{\xx}(t_i)\}_{i=1}^n$ to obtain a continuous-time estimate of robot
trajectory~\ie~$\hat{\xx}_\text{sp} : \Wcal_k \to
\mathrm{SE}(3)$ such
that $\hat{\xx}_\text{sp}(t)$ denotes the estimated pose at time $t \in \Wcal_k$.
 We then perform another
cubic B-spline interpolation, this time between poses in
$\{\breve{\xx}(t_i)\}_{i=1}^{n}$ where $\breve{\xx}(t_i)$ is an estimate of
robot pose at time $t_i$ obtained by linearly interpolating our latest estimate
of robot pose at the two closest timestamps in $\Timu_k$. This gives another
continuous-time estimate of robot's trajectory that we denote by
$\breve{\xx}_\text{sp} : \Wcal_k \to \mathrm{SE}(3)$. We now update our
estimate of robot's trajectory at IMU timestamps according to\footnote{In
		\eqref{eq:imuTrajUpdate}, $\hat{\xx}(t)$, $\hat{\xx}_\text{sp}(t)$,
		and $\breve{\xx}_\text{sp}(t)$ are meant to be seen as
		elements of $\mathrm{SE}(3)$ (using the natural identification between
elements of $\mathrm{SO}(3) \times \Rset^3$ and $\mathrm{SE}(3)$).}
\begin{align}
		\hat{\xx}(t) \leftarrow \hat{\xx}_\text{sp}(t) \cdot
		({\breve{\xx}_\text{sp}(t)})^{-1} \cdot \hat{\xx}(t), \quad \forall t
		\in \Timu_k.
		\label{eq:imuTrajUpdate}
\end{align}

\noindent\textbf{Step 2.3 - Updating Surfel Positions:} 
The surfels' positions in the world frame are determined by the estimated
robot trajectory. We use the updated estimate of the robot's trajectory at IMU timestamps
$\{\hat{\xx}(t) : t \in \Timu_k \}$ to reproject surfels generated in $\Wcal_k$ in the world frame.
Note that this step may result in a new set of surfel correspondences in
the next correspondence step (Section~\ref{sec:correspondence}).

\subsection{Cost Functions}
\label{sec:costs}
\begin{figure}[t]
\centering
\includegraphics[width=1.0\linewidth]{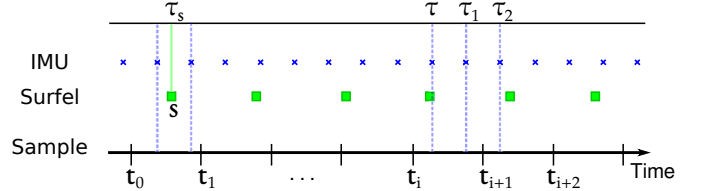}
\caption{\small{A schematic of Wildcat data fusion procedure. The sample poses, whose timestamps ($t_i$) fall between IMU timestamps, are initialised with linear interpolation between two closest IMU poses. Similarly, surfels,~\eg~$s$, are initialised based on linear interpolation between IMU poses.}}
\label{fig:data_fusion}
\end{figure}

In this section, we introduce the cost functions used in \eqref{eq:optproblem} for optimising
robot's trajectory. Recall that our optimisation variables are pose corrections 
$\{(\Rc_i,\pc_i)\}_{i=1}^{n}$ computed to correct the estimated robot poses at
the \emph{sampling
times} $\{t_i\}_{i=1}^{n}$. The sampling times, however, are not aligned with surfel or IMU timestamps. As shown in~\figref{fig:data_fusion},
we use linear interpolation (on $\mathfrak{so}(3) \times
\Rset^3$) to relate the IMU measurements and estimated surfels' positions to 
correction poses. Specifically, consider an arbitrary
time $\tau \in \Wcal_k$ where $\tau$ is not necessarily in $\{t_i\}_{i=1}^{n}$. Let
$t_{a},t_{b} \in \{t_i\}_{i=1}^{n}$ be the two closest sampling
times to $\tau$ such that $t_{a} \leq \tau \leq t_{b}$ and define $\alpha_\tau
\triangleq (\tau - t_a)/(t_b - t_a)$.  We then denote the \emph{interpolated correction pose}
at time $\tau$ by $(\bar{\RR}_{\tau}^{\text{cor}},\bar{\pp}_{\tau}^{\text{cor}})
\in \mathrm{SO}(3) \times \Rset^{3}$ where
\begin{align}
		\bar{\RR}_{\tau}^{\text{cor}} & \triangleq
		\rotlinterp(\rc_{a},\rc_{b},\alpha_\tau)
		\label{eq:interpolatedCorrectionRot},
		\\
		\bar{\pp}_{\tau}^{\text{cor}} & \triangleq 
		\linterp(\pc_{a},\pc_{b},\alpha_\tau).
		\label{eq:interpolatedCorrectionTrans}
\end{align}
Here $\linterp$ denotes the linear interpolation operator  \eqref{eq:linearInterp}, and $\rotlinterp$ interpolates rotations in a similar fashion.\footnote{Interpolating rotations can be done in multiple ways such using spherical linear interpolation (Slerp) for unit quaternions or by following a geodesic on $\mathrm{SO}(3)$ between the two rotation matrices for duration of $\alpha_\tau$. Our current implementation uses an approximation of Slerp.}
With this notation, we are now ready to describe the cost functions below.

\noindent\textbf{Surfel-Matching Cost Functions:}
Consider a pair of matched surfels $(s,s') \in \Mcal$. 
We use interpolated correction poses at surfel times $\tau_s$ and
$\tau_{s'}$ to formulate a point-to-plane-type cost function
\cite{segal2009generalized} that penalises misalignment between $s$ and $s'$
\emph{after} applying correction poses.
The eigenvector corresponding to the smallest eigenvalue of the
combined sample covariance matrices is used as our estimate for the normal vector ${\nn}_{s,s'}$ to the planar patch
captured in $s$ and $s'$ (see Section~\ref{sec:surfel}). Our point-to-plane-type
cost function is defined as
\begin{align}
		f_\text{match}^{s,s'} \triangleq w_{s,s'}
		\Big\langle 
		\nn_{s,s'}, 
		\bar{\RR}_{\tau_{s'}}^{\text{cor}}  \hat{\su}_{s'} + \bar{\pp}_{\tau_{s'}}^{\text{cor}}
		- 
		\bar{\RR}_{\tau_s}^{\text{cor}}  \hat{\su}_s - \bar{\pp}_{\tau_s}^{\text{cor}}
		\Big\rangle^2,
		\label{eq:cost_surfel}
\end{align}
where, $\hat{\su}_s$ and $\hat{\su}_{s'}$ denote the current estimate of the
positions of surfels $s$ and $s'$, respectively, and $w_{s,s'} \triangleq
{1}/({\sigma^2 + \lambda_\text{min}^{s,s'}})$ is a scalar weight defined using the
lidar noise variance $\sigma^2$ and the smallest eigenvalue
$\lambda_{\text{min}}^{s,s'}$ of the combined covariance matrix which quantifies the
thickness of the combined surfels. 

\noindent\textbf{IMU Cost Functions:}
Let ${\ac}_{\tau}$ and ${\om}_\tau$ be the linear acceleration and angular velocity
measured by IMU at time $\tau \in \Timu_k$. These measurements are modelled as,
\begin{align}
		{\ac}_\tau & = \RR(\tau)^\top \left({}_\text{w}\ac(\tau) - \grav\right) + \bb_{a}(\tau) +
		\boldsymbol\epsilon_{a}(\tau) \label{eq:linearAccModel}, \\
		{\om}_\tau &= \om(\tau) + \bb_{\omega}(\tau) +
		\boldsymbol\epsilon_{\omega}(\tau),
		\label{eq:IMUmodel}
\end{align}
where, (i) ${}_\text{w}\ac(\tau),\om(\tau) \in \Rset^3$ denote the true linear acceleration of
body in the world frame and the
angular velocity of body relative to the world frame expresses in the body
frame, respectively; (ii) IMU biases are denoted by
$\bb_a(\tau), \bb_{\omega}(\tau)\in \Rset^3$; (iii) 
$\boldsymbol\epsilon_a(\tau)$ and $\boldsymbol\epsilon_\omega(\tau)$ are white
 Gaussian noises; and (iv) $\grav$ is the gravity
vector in the world frame.

Now consider the IMU measurement received at time $\tau \in \Timu_k$ and let
$\tau_1,\tau_2$ be the timestamps of
the two subsequent IMU measurements. We have $\tau_2 \approx \tau_1 +
\Delta t_\text{imu} \approx \tau + 2\Delta t_\text{imu}$ where $\Delta t_\text{imu}$ is the (nominal) time difference between
subsequent IMU measurements (in our case, $\Delta t_\text{imu} = 0.01$s).
The IMU cost function corresponding to measurements collected at $\tau \in \Timu_k$ can be written as $f^\tau_{\text{imu}} = f_a^\tau +
f_\omega^\tau + f^\tau_\text{bias}$
where,
\begin{align}
		f_a^\tau & \triangleq 
		\left\| \om_\tau - \hat{\om}(\tau)
		- \bb_{\omega} \right\|^{2}_{\Sigma_a^{-1}}, \\
		f_\omega^\tau & \triangleq 
		\left\| \widetilde{\RR}(\tau) (\ac_\tau -\bb_{a})-
		{}_\text{w}\hat{\ac}(\tau) + \grav
		\right\|^2_{\Sigma_\omega^{-1}}, \\
		f_\text{bias}^\tau & \triangleq 
		\| \bb_{\omega} - {\hat{\bb}_\omega} \|^2_{\Sigma_{b_\omega}^{-1}} + 
		\| \bb_{a} - {\hat{\bb}_{a}} \|^2_{\Sigma_{b_a}^{-1}}.
		\label{eq:imuCosts}
\end{align}
where, $\Sigma_{a}$, $\Sigma_{\omega}$, $\Sigma_{b_a}$, and $\Sigma_{b_\omega}$ are measurement and biases covaraince matrices, $\hat{\bb}_\omega$ and $\hat{\bb}_a$ are the latest estimates of IMU biases, and $\hat{\om}(\tau)$ and ${}_\text{w}\hat{\ac}(\tau)$ are estimates of $\om(\tau)$ and ${}_\text{w}\ac(\tau)$ \emph{after} applying correction poses using the Euler's method,
\begin{align}
		\hat{\om}(\tau) & \triangleq \frac{1}{\Delta t_{\text{imu}}} \left[ \log \left( \widetilde{\RR}(\tau)^\top
		\widetilde{\RR}(\tau_1)  \right) \right]^\vee \\
		{}_\text{w}\hat{\ac}(\tau) & \triangleq \frac{1}{{\Delta
		t_\text{imu}}^2}\Big(\widetilde{\pp}(\tau_2) - 2
		\widetilde{\pp}(\tau_1) + \widetilde{\pp}(\tau) \Big),
\end{align}
in which, $\widetilde{\RR}(t) = \bar{\RR}_{t}^{\text{cor}} \, \hat{\RR}(t)$
and $ \widetilde{\pp}(t) = \bar{\pp}_{t}^{\text{cor}} +
\hat{\pp}(t)$ for $t \in \{\tau, \tau_1, \tau_2\}$ describe robot
pose at time $t$ after applying the interpolated correction poses; see \eqref{eq:interpolatedCorrectionRot} and
\eqref{eq:interpolatedCorrectionTrans}.

%% file: chapters/pgo.tex
\section{Wildcat Pose-Graph Optimisation}
\label{sec:pgo} 
In this section, we describe key components of Wildcat's pose-graph optimisation (PGO) module.
Wildcat's odometry module estimates robot's trajectory only using \emph{local}
information and thus inevitably suffers from accumulation of error over time.
The PGO module addresses this issue by optimising trajectory
using \emph{global} information, albeit at a lower temporal resolution.

\subsection{Submap Generation} 
\label{sec:frame}
The building blocks of our PGO module are \emph{submaps}. Submaps encapsulate data over a \emph{short} fixed-length time window. Specifically, each submap is a six-second
bundle of odometry estimates, accumulated local surfel
map, and an estimate of the direction of the gravity vector in the submap's local
coordinate frame.\footnote{This estimate is obtained using accelerometer
measurements and the trajectory estimated by odometry; see
\eqref{eq:linearAccModel}.} Wildcat generates submaps periodically (\eg~every five seconds) after the odometry
module
fully processes the corresponding time interval. The error accumulated
\emph{within} a submap is negligible because each submap's internal structure is
already optimised by the odometry's sliding-window optimisation scheme. This allows the PGO
module to treat each submap as a \emph{rigid} block whose configuration in the
world coordinate frame can be represented by that of its \emph{local} coordinate
frame. 

In multi-agent collaborative SLAM scenarios, each agent synchronises its database of
submaps (containing submaps generated by itself and others) with other
agents (\ie~other robots or the base station) within its communication range via peer-to-peer
communication. We refer the reader to \cite{hudson2021heterogeneous} for
additional information about our ROS-based data sharing system, Mule. The
maximum size of each submap with a lidar range of 100 m is about 500 KB,
whereas the
average submap size in underground SubT events was about 100-170 KB.
\cite{hudson2021heterogeneous}. Therefore, Wildcat can easily share submaps between
the agents with a modest communication bandwidth.

\subsection{Pose Graph}
Recall that nodes in a pose graph represent (unknown) poses and edges represent relative
noisy rigid-body transformations between the corresponding pairs of poses.
\subsubsection{Nodes}
The nodes in our pose graph initially correspond to Wildcat submaps. More specifically,
each node represents the pose of a submap's local coordinate frame with respect to the world coordinate
frame. Upon adding a new edge to the pose graph (see below), the PGO module merges nodes whose corresponding local surfel maps have
significant overlap and whose Mahalanobis distance is below a threshold relative to a single node. By merging redundant nodes, the computational cost of
our PGO module grows with the size of explored environment rather than mission
duration.

\subsubsection{Edges}
There are two types of edges in a pose graph, namely odometry and loop-closure edges.
The odometry edges connect consecutive nodes and are obtained from 
the odometry module's estimate of relative rigid-body transformation between the corresponding
two nodes. By contrast, loop-closure edges (typically) link non-consecutive nodes and are
computed by aligning the local maps of the corresponding pairs of nodes.
If the overlap between the corresponding submaps is sufficient and if the uncertainty associated to their relative pose is below a threshold, we use point-to-plane Iterative Closest Point (ICP) to align the surfel submaps. Otherwise, we first obtain a rough alignment using global methods such as \cite{makadia2006fully} and use that to initialise ICP.

Potential loop closure candidates are detected either based on a Mahalanobis distance search radius or by using existing place recognition methods. The PGO modular design allows us to easily integrate place recognition techniques such as Scan Context \cite{gkim-2018-iros} with Wildcat.
In either case, the loop closure candidate is added to the pose graph when it
passes a gating test based on the Mahalanobis distance.

\subsection{Optimisation}
We denote the pose graph with $\mathcal{G} = (\mathcal{V},\mathcal{E})$ where
$\mathcal{V} = [m]$ represents pose graph
nodes and $\mathcal{E}$ is the pose graph edge set.
Let $\xx_i = (\RR_i,\pp_i) \in \mathrm{SO}(3) \times \Rset^3$ denote the pose of
the $i$th pose graph node in the world coordinate frame.
The standard cost function minimised by pose-graph optimisation methods can be written as
\begin{align}
		f_\text{pgo}(\xx_1,\ldots,\xx_m) = \sum_{(i,j) \in \mathcal{E}}
		f_{ij}(\xx_i,\xx_j),
		\label{eq:pgoStandardCost}
\end{align}
where $f_{ij}(\xx_i,\xx_j)$ is the standard squared error residual for the relative
rigid-body transformation between $\xx_i$ and $\xx_j$.

We add an extra term to the standard PGO cost
function \eqref{eq:pgoStandardCost} to leverage information available about the
vertical direction (direction of gravity in the world frame) through accelerometer
measurements. Specifically, our
PGO module minimises the following cost function,
\begin{align}
		f_\text{pgo}(\xx_1,\ldots,\xx_m) +
		f_\text{up}(\RR_1,\ldots,\RR_m),
		\label{eq:ourPGOCost}
\end{align}
in which,
\begin{align}
		f_{\text{up}}(\RR_1,\ldots,\RR_m) \triangleq \sum_{i \in \mathcal{V}} \| \RR_i \hat{\mathbf{u}}_i -
		{}_\text{w}{\mathbf{u}} \|^2
\end{align}
where, $\hat{\mathbf{u}}_i$ is the estimated vertical direction (\ie~$\|\hat{\mathbf{u}}_i \| = 1$) in the local
frame of the $i$th node and ${}_\text{w}\mathbf{u} \triangleq [0 \,\, 0 \,\, 1]^\top$ is the vertical
direction in the world frame. As we mentioned earlier in
Section~\ref{sec:frame}, $\hat{\mathbf{u}}_i$ is calculated using
\eqref{eq:linearAccModel} and odometry module's estimated robot trajectory at
IMU timestamps.
Similar to the odometry module, to be robust to outliers (\eg~false-positive
loop closures) the PGO module minimises \eqref{eq:ourPGOCost}
using an IRLS scheme based on the Cauchy M-estimator.

%% file: chapters/experiments.tex
\section{Experiments}
\label{sec:experiments}

In this section, we experimentally evaluate Wildcat on a diverse collection of real datasets and compare its results with two state-of-the-art lidar-inertial SLAM methods, namely FAST-LIO2~\cite{xu2021fast2} and LIO-SAM~\cite{shan2020lio}.

\subsection{Summary of Datasets}
The datasets used in our experimental analysis are as follows (see also Table~\ref{tab:datasets} for a summary).
\subsubsection{DARPA SubT Dataset} 
\begin{figure}[t]
\centering
\includegraphics[width=0.9\linewidth]{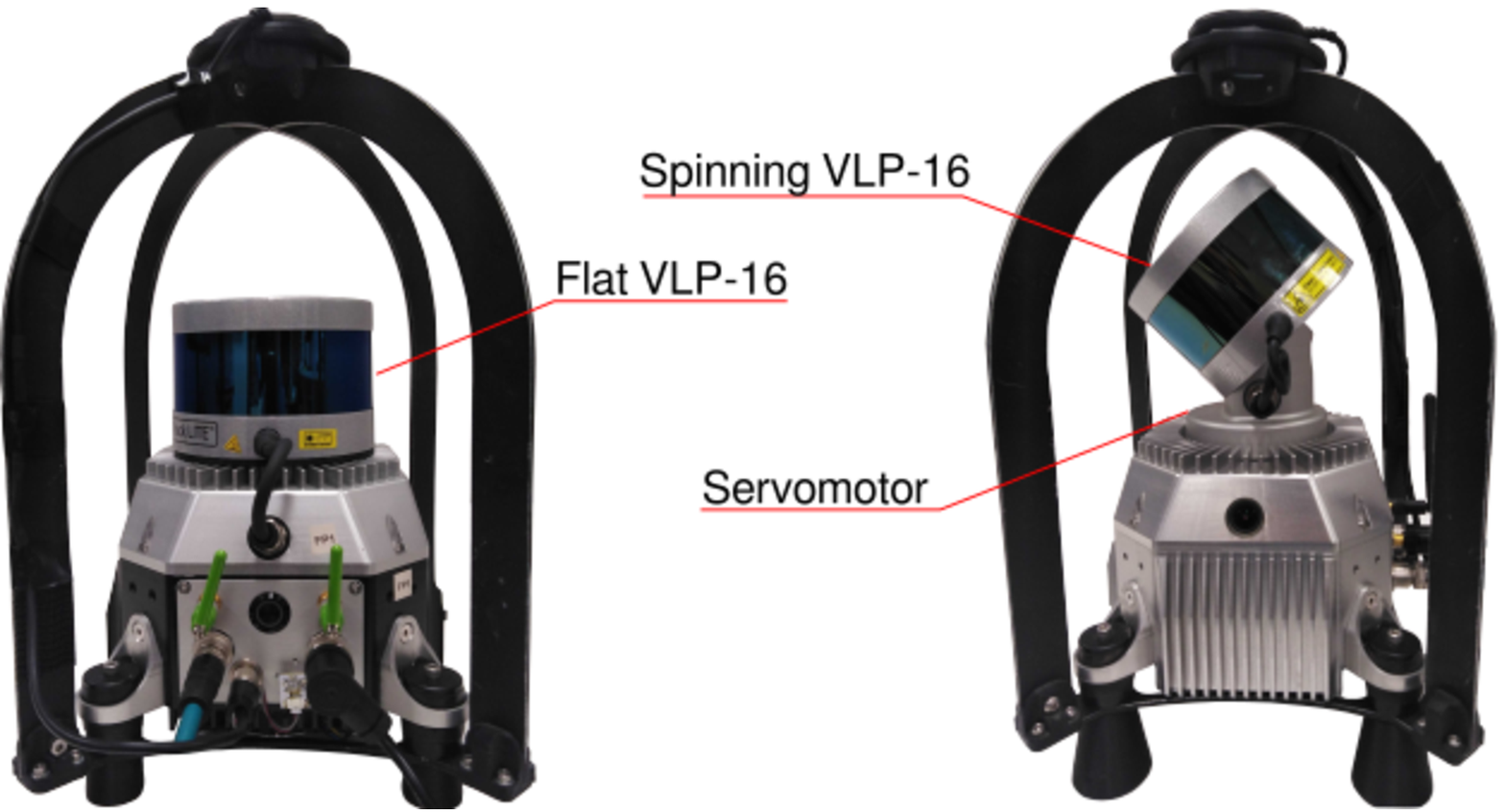}
\caption{\small{\flatpack{} (left) versus \spinningpack{} (right).}}
\label{fig:packs_photos}
\end{figure}

This dataset was collected by Team CSIRO Data61 comprises of two Boston Dynamics Spot robots and two BIA5 ATR tracked robots at the SubT Final Event in Louisville Mega Cavern. Each robot was equipped with a spinning pack, designed and engineered at CSIRO. A picture of this pack is shown in~\figref{fig:packs_photos} (right). The spinning pack (hereafter referred to as \spinningpack{}) is composed of a Velodyne VLP-16, with the measurement rate set to 20 Hz, a 9-DoF 3DM-CV5 IMU measuring angular velocity and linear acceleration at 100 Hz, and four RGB cameras. In the \spinningpack{}, the Velodyne lidar is mounted at an inclined angle on a servomotor spinning around the sensor's $z$ axis at 0.5 Hz. The servomotor is designed in a way that the spinning Velodyne VLP-16 provides 120\degree~vertical Field of View (FoV).
We use the ground truth point cloud map provided by DARPA to evaluate Wildcat's multi-agent mapping accuracy. 

 \subsubsection{MulRan Dataset} We use the DCC03 sequence of the MulRan~\cite{Kim2020MulRanMR} dataset. This publicly available urban driving dataset was collected using an Ouster OS1-64 (at 10 Hz with a range of 120 m) on a vehicle in Daejeon, South Korea. The length of this sequence is about 5 km. Combining GPS, fiber optic gyro and SLAM, MulRan provides the vehicle motion ground truth in 6 DoF at 100 Hz.

\begin{figure}[t]
\centering
\includegraphics[width=1.0\linewidth]{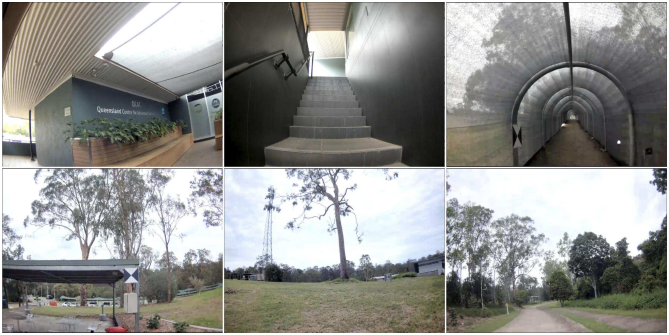}
\caption{\small{Photos from the QCAT (\spinningpack) dataset indoors and outdoors. The QCAT dataset includes various types of environments. The confined areas such as through the tunnel or stair cases challenge SLAM due to the degeneracy of these areas. In some of the photos, surveyed targets can be seen.}}
\label{fig:qcat_photos}
\end{figure}

\begin{table*}[]
\small
\centering
\caption{Description of the datasets used for experimental evaluation of Wildcat.}
\label{tab:datasets}
\begin{tabular}{@{}lccccc@{}}
\toprule
Dataset              & Description                                                                         & Ground Truth           & Lidar           & Spinning & No. of Agents                                                                   \\ \midrule
DARPA SubT Final Event    &  Subterranean Environments                                                                              & DARPA's Pointlcoud Map &  VLP-16 & \cmark                 &  4 \\
MulRan DCC03 \cite{Kim2020MulRanMR}       & Urban Driving/Outdoor & GPS/INS/SLAM                    &  OS1-64          & \xmark            & 1                                                                           \\
QCAT (\flatpack{})     &   Hand-held Platform/Indoor/Outdoor                                                                                  & Surveyed Targets    &  VLP-16 & \xmark            & 1                                                                          \\
QCAT (\spinningpack{}) &  Hand-held Platform/Indoor/Outdoor                                                                                   & Surveyed Targets     &  VLP-16 & \cmark               & 1                                                                             \\ \bottomrule
\end{tabular}
\end{table*}

\subsubsection{QCAT Dataset}
This in-house dataset, including two large scale sequences named QCAT (\flatpack{}) and QCAT (\spinningpack{}), has been collected at the Queensland Centre for Advanced Technologies (QCAT) in Brisbane, Australia. 
These sequences were captured by two hand-held perception packs, a \flatpack{} and a \spinningpack{}, both designed at CSIRO. In contrast to the \spinningpack{}, described earlier, the FoV in the \flatpack{} is equal to the vertical FoV provided by VLP-16 (~\ie~30\degree). ~\figref{fig:packs_photos} (left) shows a picture of the flat pack.
For a fair comparison between~\flatpack~and~\spinningpack, each sequence was collected roughly through the same path across QCAT with the duration of about 2 hours each at a walking speed. The traverse distance for each dataset is about 5 km.~\figref{fig:qcat_photos} shows several photos of the dataset across QCAT.
The QCAT dataset is uniquely diverse and challenging and due to travelling indoors and outdoors,  providing an accurate pose ground truth is not feasible in such a complex and diverse environment. Instead, we deployed and surveyed over 60 targets scattered across the site (\figref{fig:qcat_data_fp1}). This ground truth, described in~\secref{sec:map}, enables us to evaluate the mapping accuracy of SLAM systems.

\subsection{Results on DARPA SubT Final Event}
\label{sec:darpa}

\begin{figure}[t]
\centering
\subfloat{\includegraphics[width=1\linewidth]{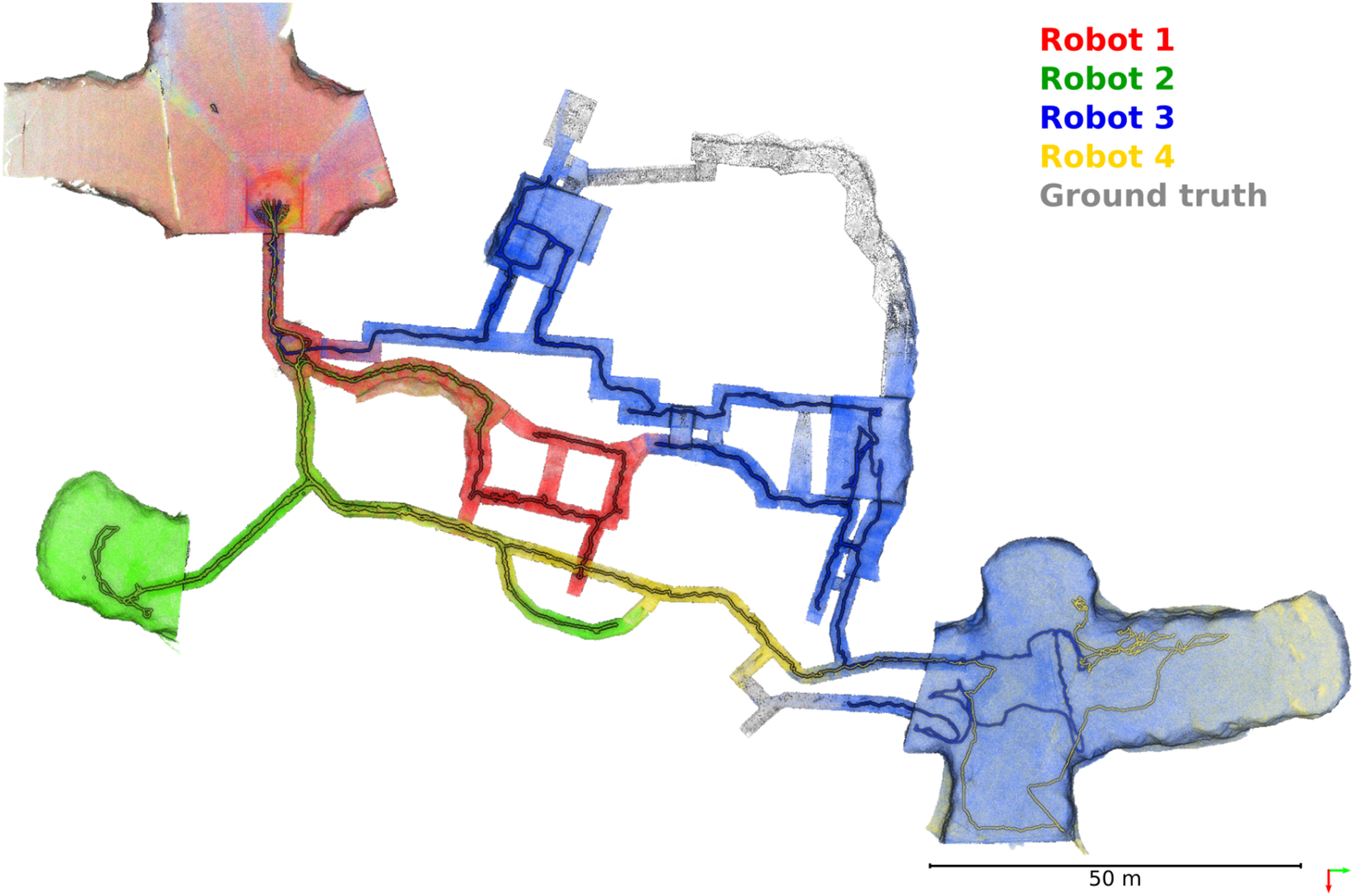}}\\
    \subfloat{\includegraphics[width=1\linewidth]{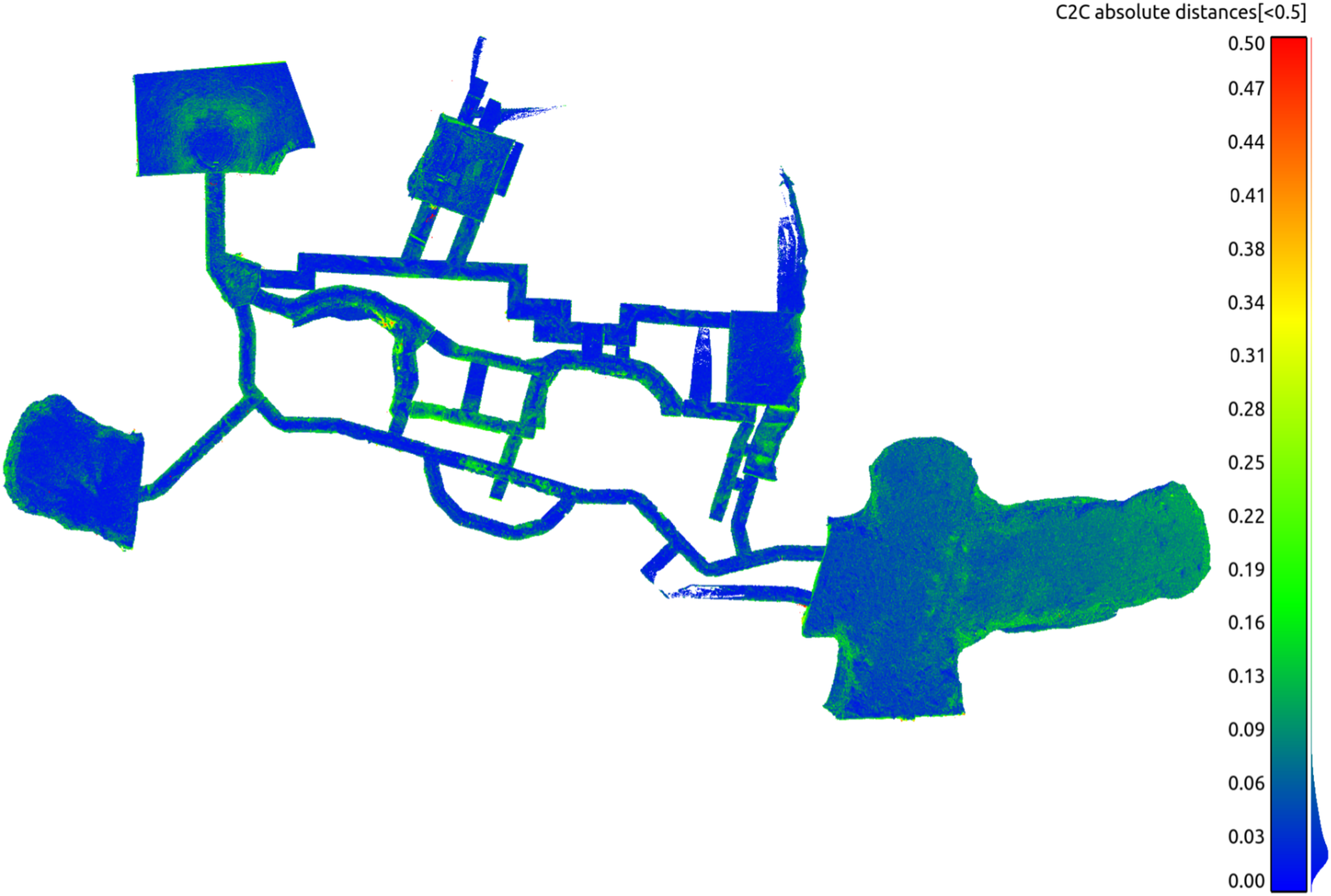}}%

\caption{\small{\textbf{Top}: Multi-agent globally optimised Wildcat SLAM map from the robots deployed during the prize run at the DARPA Subterranean Challenge Final Event at the Louisville Mega Cavern, KY. The multi-agent map, shown in red, blue, green and yellow, was reported by DARPA to have ``0\% deviation" from the surveyed ground truth (in grey) and cover 91\% of the course. Note that the clouds were decimated and filtered to remove noisy points caused by interference between lidar sensors at long ranges. \textbf{Bottom}: Point-wise comparison between the map generated on the fly and the ground truth map provided by DARPA.}}
\label{fig:darpa}
\end{figure}

\begin{figure}[t]
\centering
\includegraphics[width=0.9\linewidth]{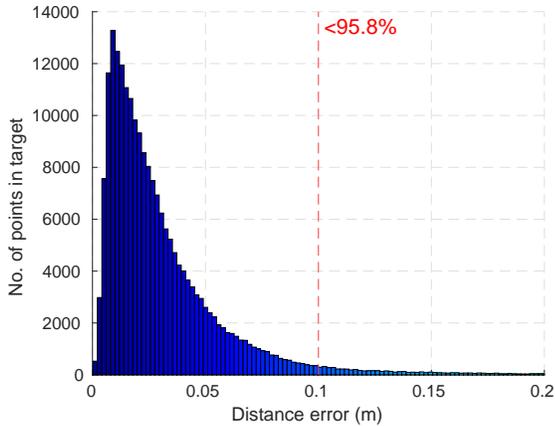}
\caption{\small{Histogram of point-wise comparison between the Wildcat map generated on-the-fly between multiple agents and the high-resolution ground truth.}}
\label{fig:darpa_hist_c2c}
\end{figure}

\figref{fig:darpa} illustrates Wildcat results for the prize run at the DARPA SubT Challenge Final Event. In this run, four robots using \spinningpack{s} started from the same area and explored a perceptually challenging subterranean environment (with sensing degradation due to dust and smoke). The robots shared their submaps with other agents via peer-to-peer communication. 
As shown in the figure, Wildcat produced a globally consistent map online which precisely aligns with DARPA's ground truth.  
According to DARPA, Wildcat's map had ``0\% deviation" from the ground truth. It is worth noting that the ground truth was generated by spending 100 person-hours using a survey-grade laser scanner, according to DARPA.

We also conduct our own point-wise comparison between the Wildcat map and the ground truth. We compare the voxelised Wildcat map (target) with a resolution of 40 cm against the ground truth map (reference) with a higher resolution of 1 cm. After a fine alignment between the two point clouds, each point in target point cloud is associated to the nearest point in reference point cloud and the distance between the corresponding points are computed.
The average distance error between the corresponding points is 3 cm with the standard deviation of 5 cm as shown in~\figref{fig:darpa} (bottom). The histogram, shown in~\figref{fig:darpa_hist_c2c}, also demonstrates that more than 95\% of the corresponding points' distances are less than 10 cm, which is consistent with DARPA's evaluation.

\subsection{Results on MulRan Dataset}
\label{sec:traj}
We use the MulRan dataset to evaluate the accuracy of odometry and SLAM trajectory estimates as it provides 6-DoF ground truth. Our evaluation is based on Relative Pose Error (RPE) for odometry evaluation and Absolute Pose Error (APE) for the evaluation of SLAM trajectory. Both metrics are computed using evo \cite{grupp2017evo}.

To run LIO-SAM on MulRan DCC03, we used the default parameters recommended by their authors. Mainly, the voxel filter parameters~\ie~\inlinecode{odometrySurfLeafSize}, \inlinecode{mappingCornerLeafSize}, \inlinecode{mappingSurfLeafSize}  were set to default as suggested for outdoor environments. For the SLAM result, the loop-closure module was enabled with the frequency set to 1~Hz to regulate loop-closure constraints. For FAST-LIO2 in which raw lidar data are used for map registration, all the parameters were set to defaults except \inlinecode{cube\_side\_length} which was set to 5000 to be compatible with the environment size in DCC03. Also, for a fair comparison between the SLAM trajectory results of Wildcat and LIO-SAM, we disabled the place recognition module in Wildcat so that the loop-closure detection is done based on a fixed search radius similar to LIO-SAM. That said, note that the search radius in Wildcat is based on the Mahalanobis distance, whereas LIO-SAM uses the Euclidean distance.

\begin{figure}[t]
    \centering
    \subfloat[RPE w.r.t the translation part]{\includegraphics[width=1.0\linewidth]{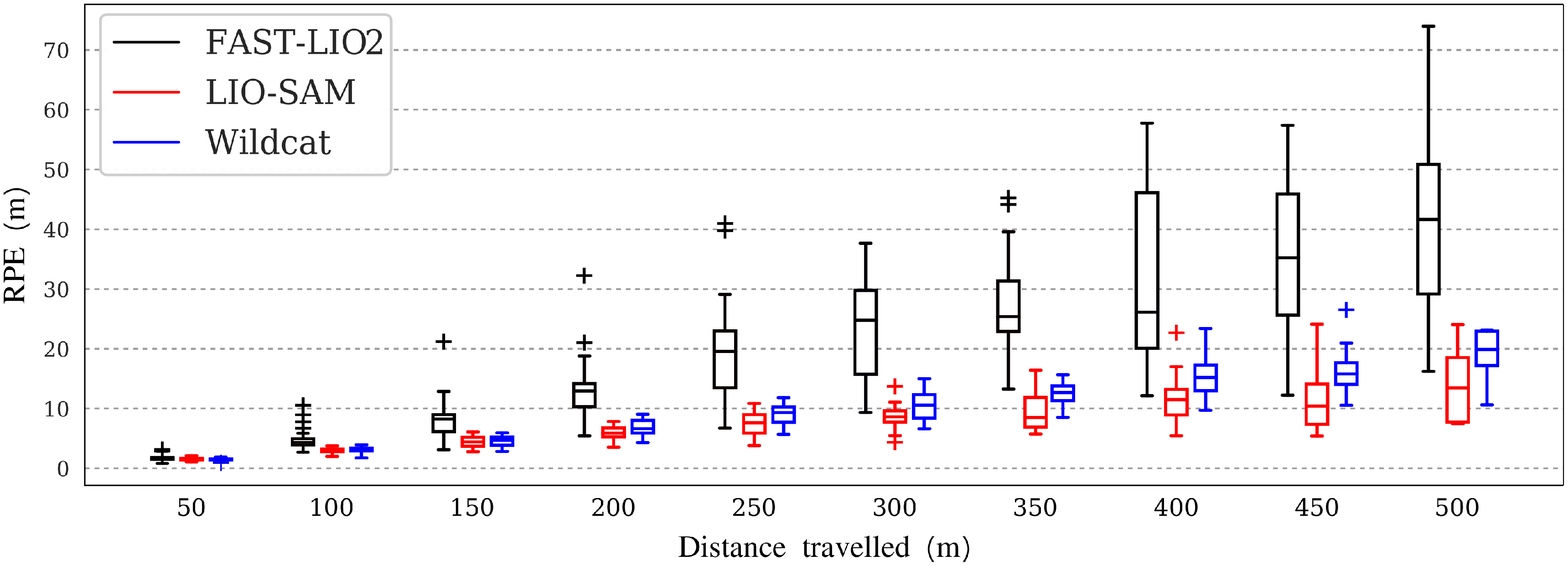}}
    \\
    \subfloat[RPE w.r.t the orientation part]{\includegraphics[width=1.0\linewidth]{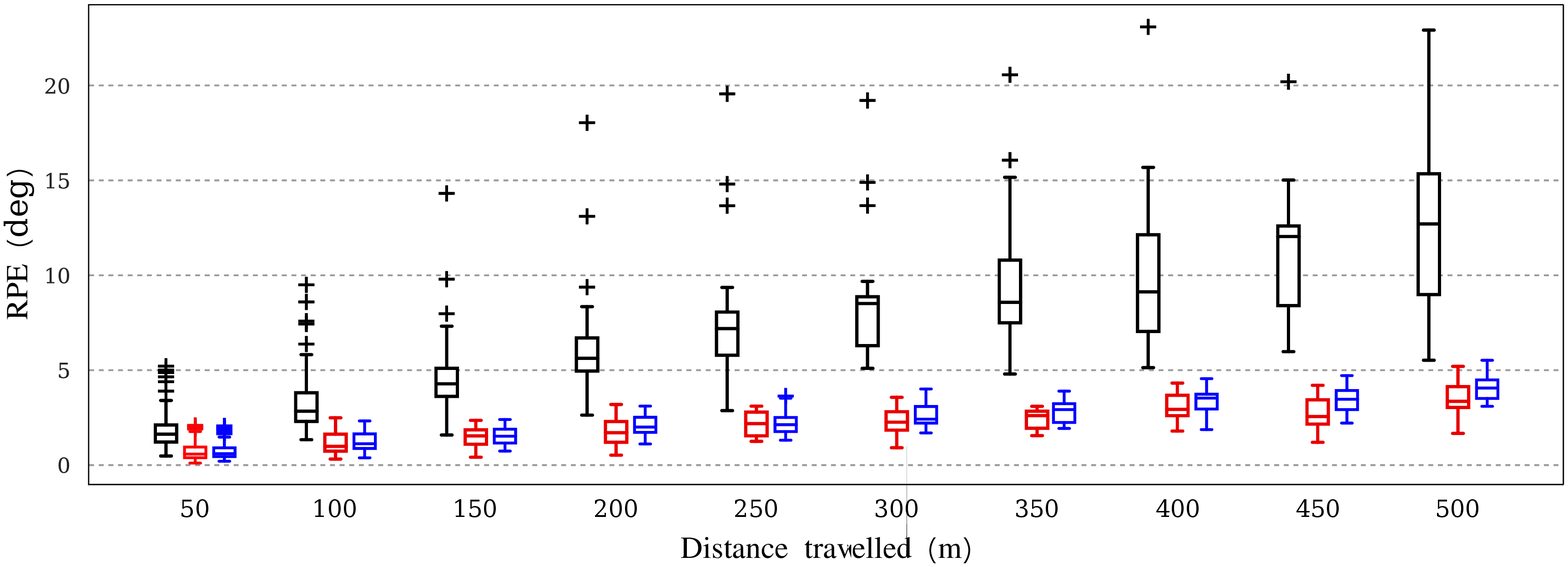}}%
    \\
    \subfloat[APE w.r.t the translation part]{\includegraphics[width=0.45\linewidth]{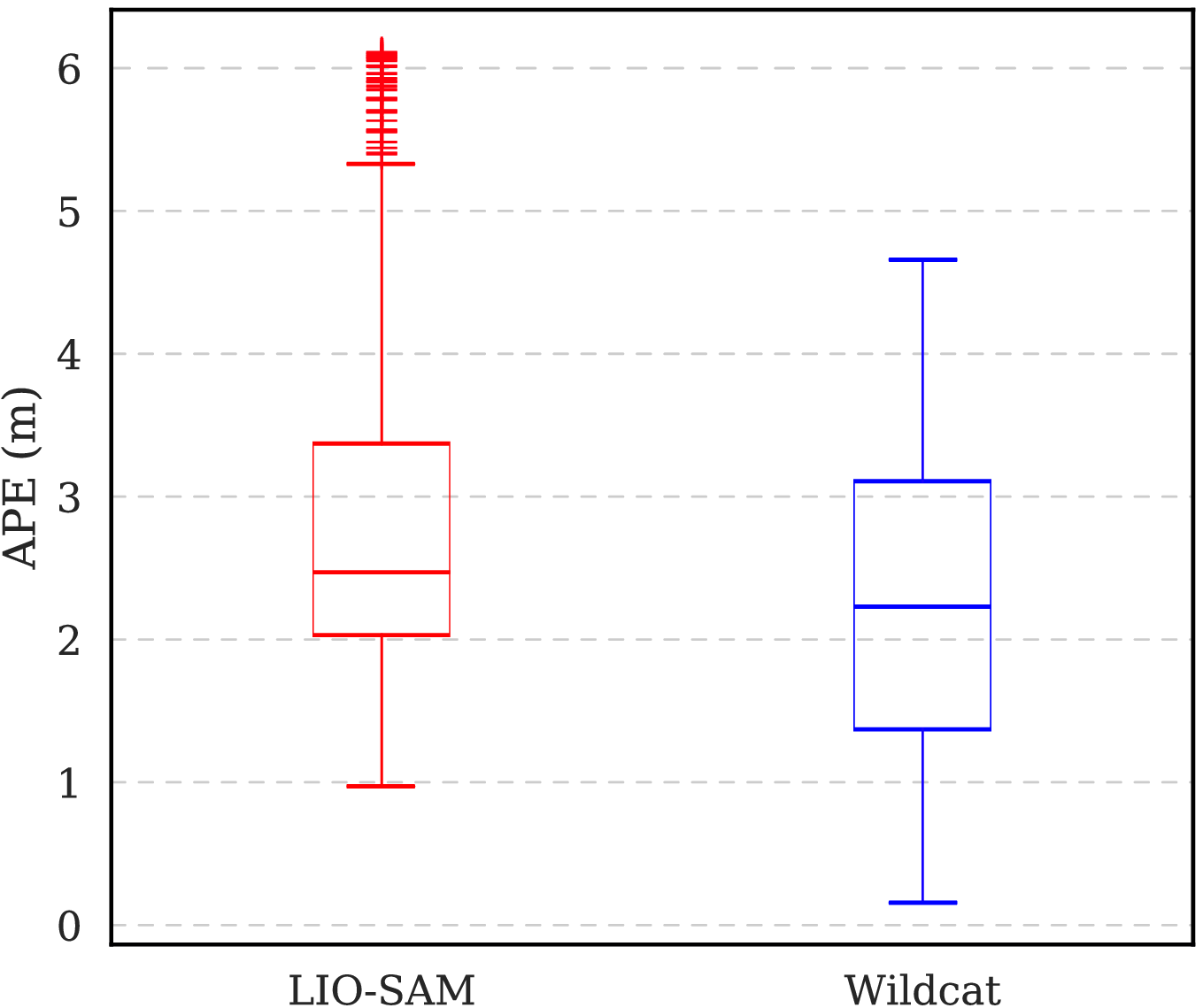}}%
    \qquad
    \subfloat[APE w.r.t the orientation part]{\includegraphics[width=0.45\linewidth]{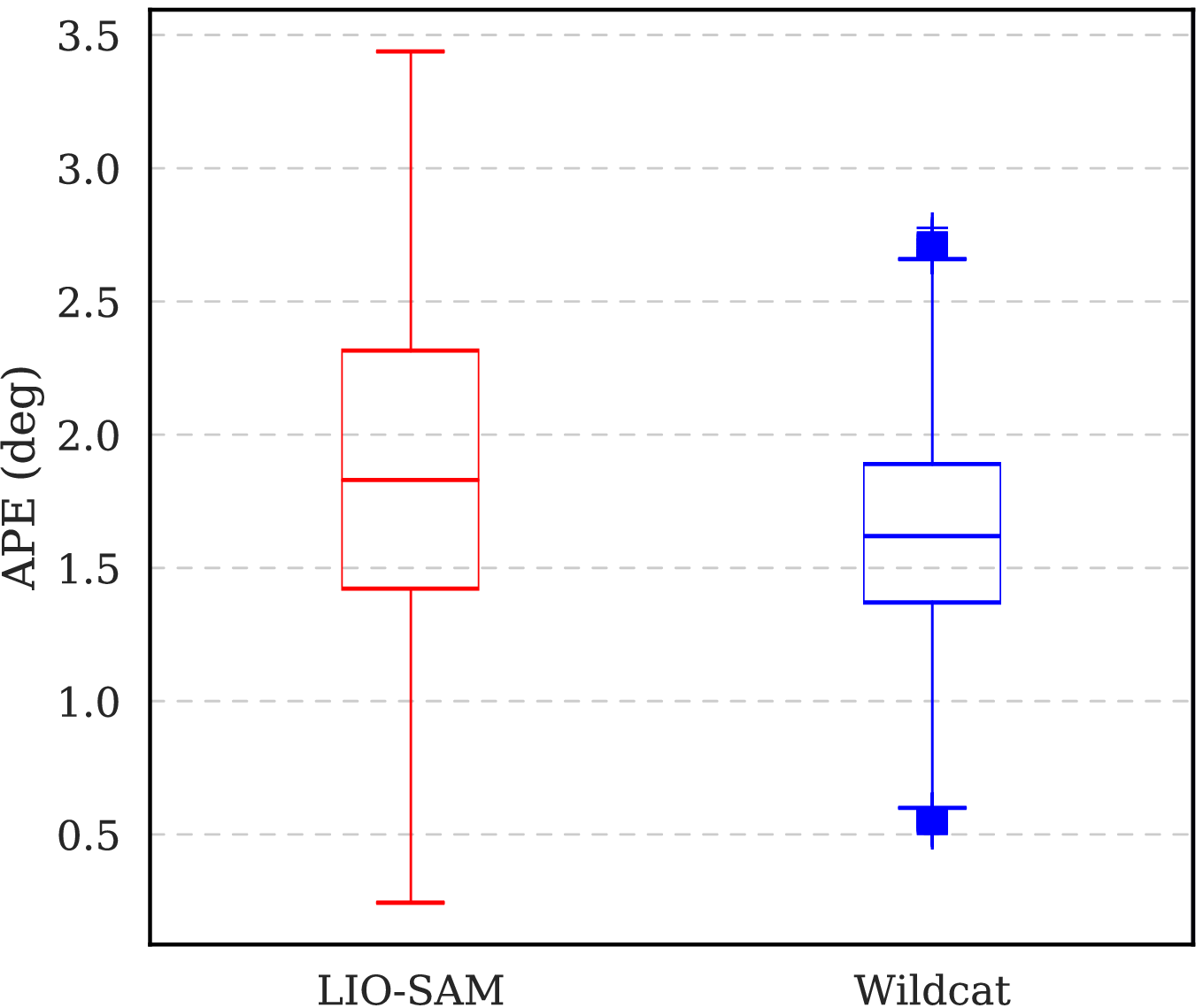}}%
    \caption{\small{Error evaluation of odometry and SLAM trajectories over MulRan DCC03. Since FAST-LIO2 is not designed to deal with drift by detecting loop-closures, we avoided to include it's results in APE evaluation.}}%
    \label{fig:rep_ape}%
\end{figure}

\figref{fig:rep_ape} shows the box plots related to RPE for the translation (a) and rotation (b) components as a function of trajectory length (varying from 50 m to 500 m).
The odometry estimates for LIO-SAM are obtained by disabling loop-closure detection.
The odometry drift of LIO-SAM is slightly smaller than Wildcat in both translation and orientation. On average, the translation drift for Wildcat and LIO-SAM odometry over this sequence is 2.9\% and 2.4\%, respectively. The average rotation error per meter of traversed trajectory for Wildcat and LIO-SAM odometry is 0.01 deg/m and 0.009 deg/m, respectively. FAST-LIO2's accuracy is worse than the other two methods and its drift grows at a higher rate. On average, the translation drift of FAST-LIO2 is 6.8\%, and its average rotation error per meter is 0.03 deg/m.

\begin{figure}[t]
\centering
\includegraphics[width=1.0\linewidth]{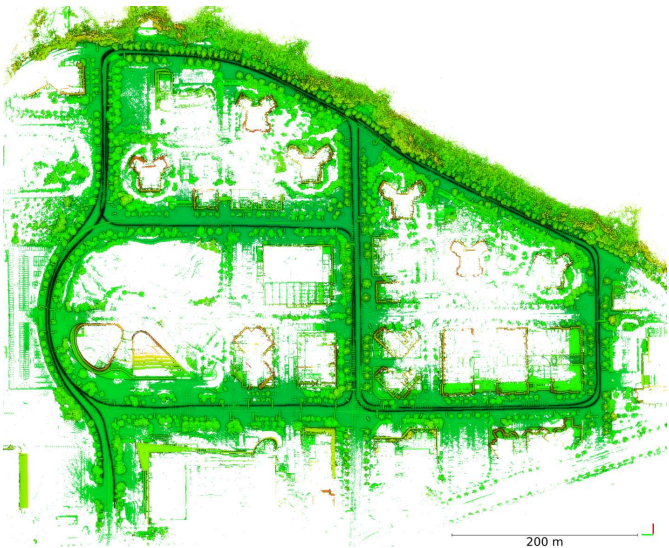}
\caption{\small{Generated Wildcat Map overlaid with the estimated trajectory over MulRan dataset sequence DCC03.}}
\label{fig:wildcat_mulran}
\end{figure}

Box plots in \figref{fig:rep_ape} (c) and (d) depict APE for translation and rotation, respectively. These results show that Wildcat achieves slightly higher accuracy performance than LIO-SAM after enabling loop closure.
This can be attributed to a number of differences between the two methods such as Wildcat PGO's gravity alignment constraints and its candidate loop-closure verification steps (see \secref{sec:pgo}).
Finally, the estimated trajectory and map by Wildcat's PGO module are shown in \figref{fig:wildcat_mulran}.

\begin{figure}[b]
\centering
\includegraphics[width=1.0\linewidth]{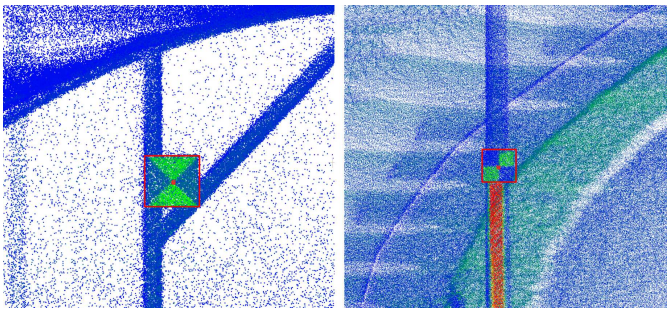}
\caption{\small{ Examples of determining the centre of targets. We select each target's centre by casting two rays from two angles of view. The target's centre is precisely computed via triangulation.}}
\label{fig:target_selection}
\end{figure}

\begin{figure}[t]
\centering
\includegraphics[width=1.0\linewidth]{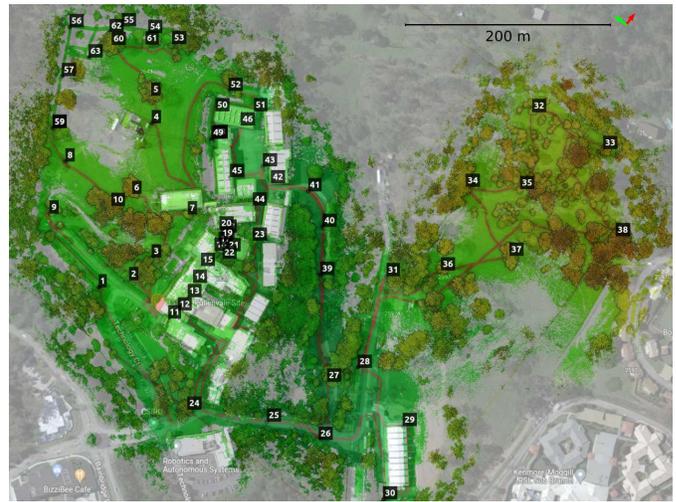}
\caption{\small{A bird's-eye view of the QCAT map, generated from the~\flatpack{} dataset along with the trajectory estimated by Wildcat, overlaid with the location of surveyed targets deployed across the site.}}
\label{fig:qcat_data_fp1}
\end{figure}

\subsection{Results on QCAT Dataset}
\label{sec:map}

We manually identified and marked the position of the centre of a number of surveyed targets in the 3D maps generated by Wildcat, LIO-SAM, and FAST-LIO2; see \figref{fig:target_selection}. These targets are placed across our campus, in various types of environments including indoor
office environments, outdoor open and forested regions as shown in~\figref{fig:qcat_data_fp1}. This variety enables us to compare the robustness and versatility of Wildcat with prior state-of-the-art methods. It is worth noting that the targets 11-21 are located in a 3-storey office environment accessed via internal stairs and the targets 53-60 are placed throughout a mock-up tunnel. These sections present a challenge to lidar SLAM systems due to their complexity and restricted view. 

To accurately estimate the position of the centre of a target, we used a tool developed by automap~\cite{automap} to fly through the maps, find the target, and cast two rays toward it from sufficiently different viewpoints. This process has several advantages. First, it prevents selecting points that are not on the target plane. Secondly, it allows us to estimate position of the centre of targets even if the 3D map points are not dense enough around the centre. 
Once the targets' centre is selected in the generated 3D map, we can register these targets with the survey points (as reference) to compute the distance error between correspondences. This evaluation process varies from point cloud to point cloud comparison (as we did in~\secref{sec:darpa}), which does not precisely show map accuracy due to the nearest neighbour procedure in data association.

To obtain the best results for FAST-LIO2 and LIO-SAM, we tuned the parameters mentioned in~\secref{sec:traj} since these methods are quite sensitive to the voxel filter parameters selected for indoor or outdoor scenarios. Hence, the results reported for these methods hereafter are for the best parameters chosen by observing the behaviour of these methods for different settings after several runs. On the contrary, Wildcat uses a single common set of voxel filter parameters for all datasets analysed in this paper.

\begin{table*}[t]
\caption{\small{Accuracy evaluation over the estimated maps in the QCAT dataset. Note that in total there are 63 targets. All targets were marked in Wildcat's map for evaluation . However, the results of LIO-SAM and FAST-LIO2 are only for subsets of targets that we could detect in their maps before these approaches failed (mainly in the tunnel).}}
\label{tab:results}
\centering
\begin{tabular}{c c c c c c c c c c c c}
 \hline
 &&&\multicolumn{9}{c}{Absolute Error} \\
 \cline{4-12}
 &&&\multicolumn{4}{c}{QCAT \flatpack} && \multicolumn{4}{c}{QCAT \spinningpack} \\ 
  \cline{4-7} \cline{9-12}
  SLAM &&&mean (m) & RMSE (m)  & std (m)& \# targets & & mean (m) & RMSE (m)  & std (m) & \# targets\\  
 \hline\hline 
 LIO-SAM~\cite{shan2020lio} &&& 0.92& 1.33& 0.97 &41& & 1.69&2.52 &1.90 &38 \\
 FAST-LIO2~\cite{xu2021fast2} &&& 1.09& 1.38& 0.85 &53& & 0.43& 0.64& 0.47 &53\\
 Wildcat (ours) &&& \textbf{0.42}& \textbf{0.46}& \textbf{0.19} &\textbf{63}&& \textbf{0.34}& \textbf{0.46}& \textbf{0.31} &\textbf{63} \\
 \hline
\end{tabular}
\end{table*}

\begin{figure*}[]
    \centering
    \subfloat[Wildcat (QCAT~\flatpack)]{\includegraphics[width=0.3\linewidth]{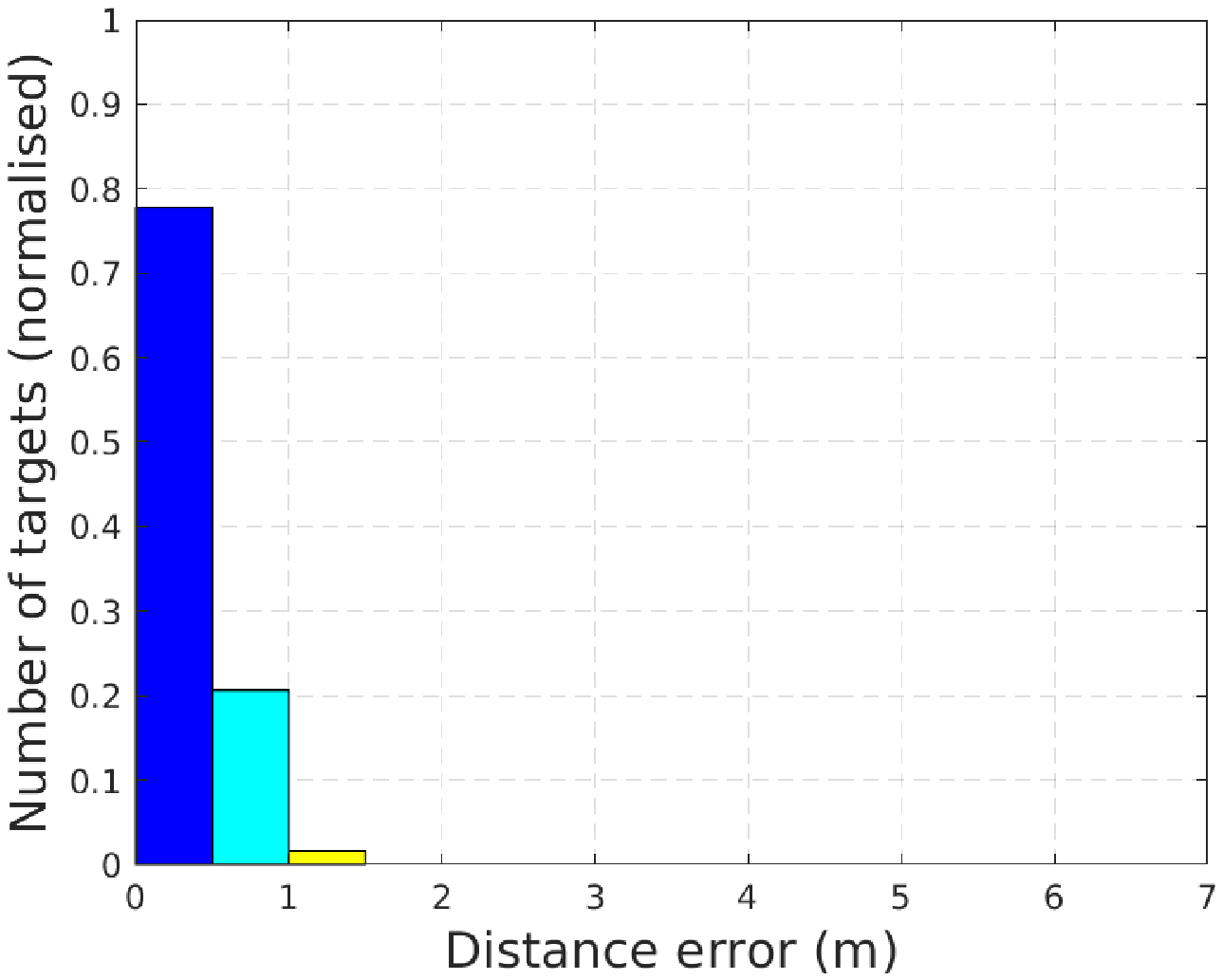}}%
    \qquad
    \subfloat[LIO-SAM (QCAT~\flatpack)]{\includegraphics[width=0.3\linewidth]{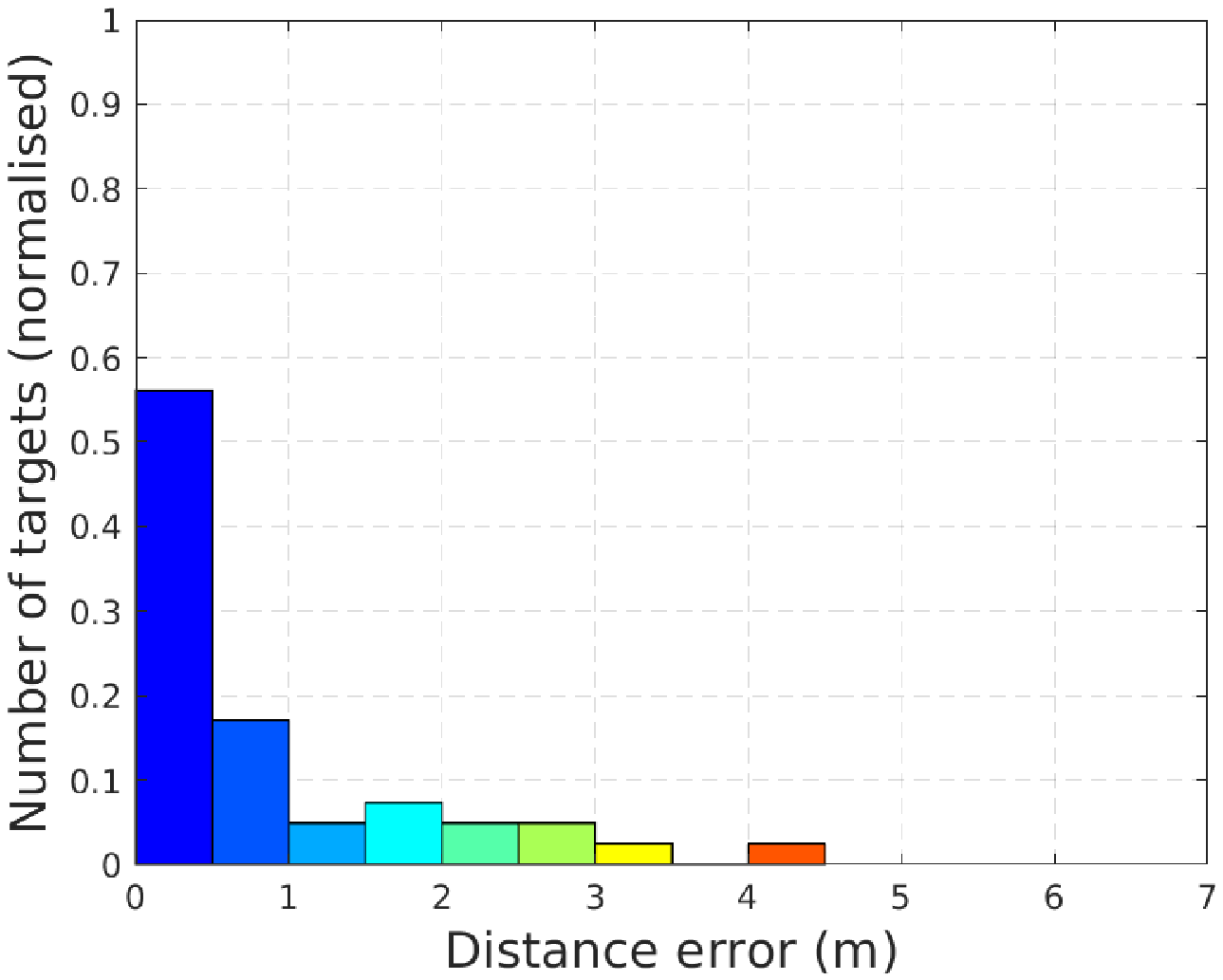}}%
    \qquad
    \subfloat[FAST-LIO2 (QCAT~\flatpack)]{\includegraphics[width=0.3\linewidth]{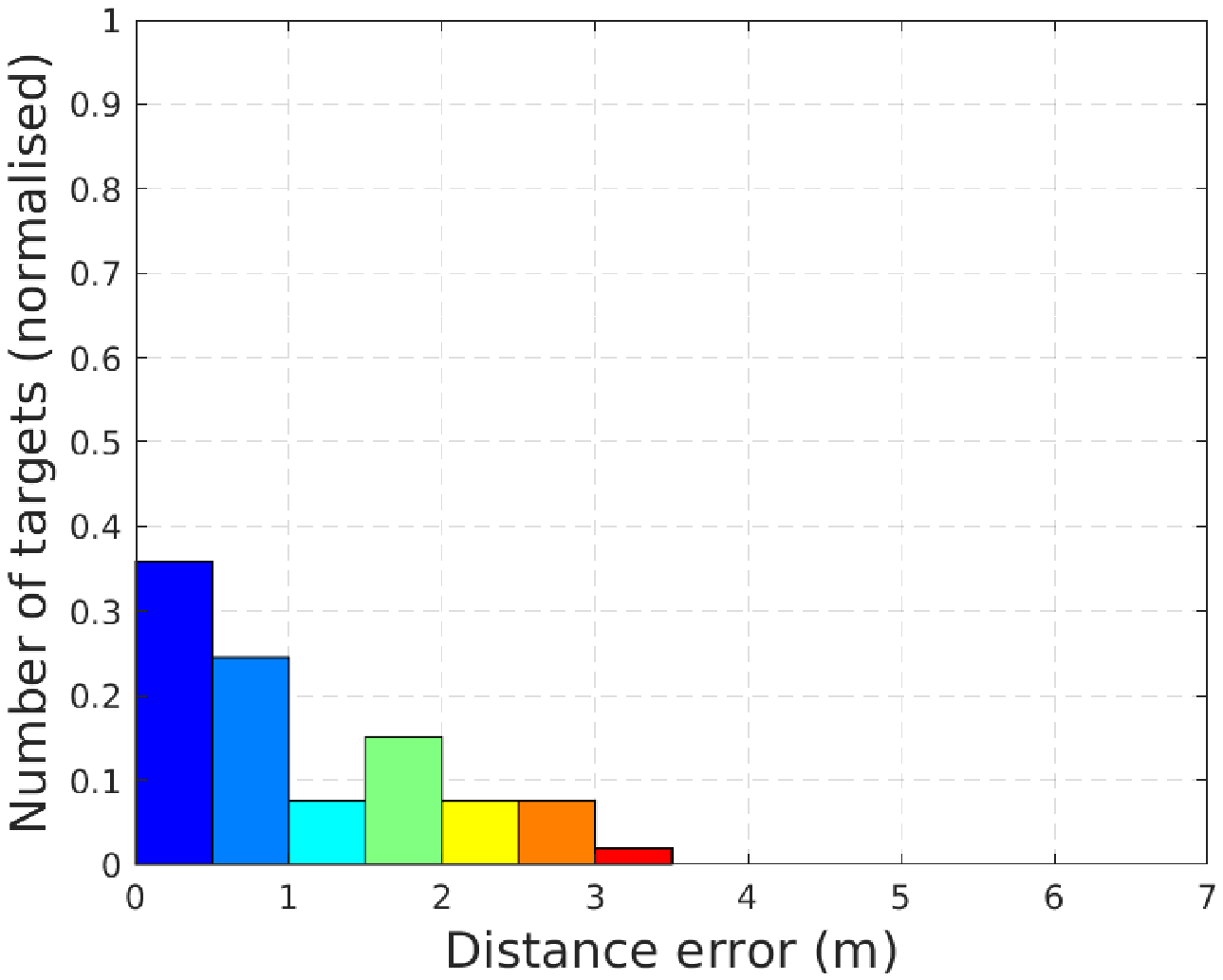}}\\
    \vspace{-2mm}
    \subfloat[Wildcat (QCAT~\spinningpack)]{\includegraphics[width=0.3\linewidth]{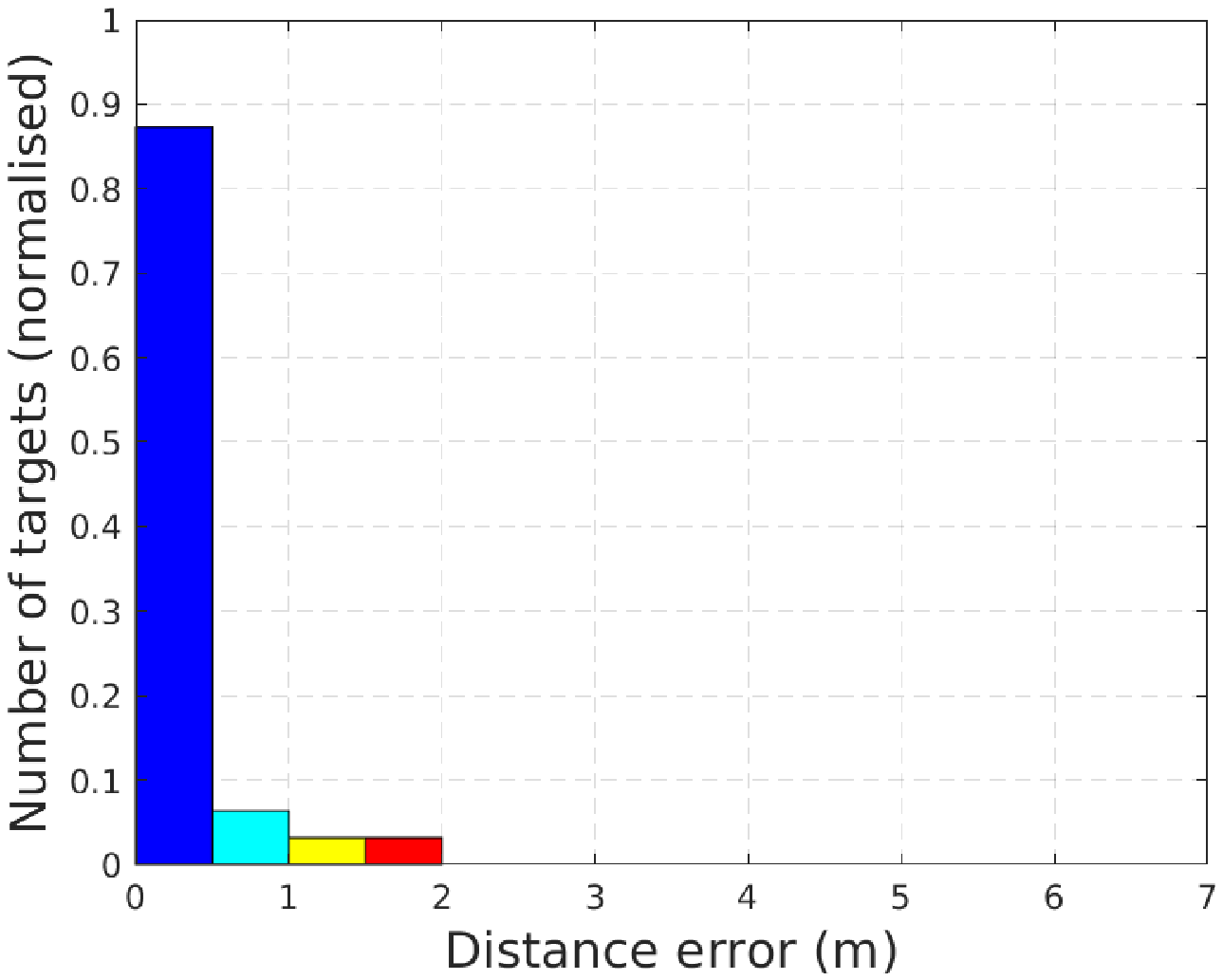}}%
    \qquad
    \subfloat[LIO-SAM (QCAT~\spinningpack)]{\includegraphics[width=0.3\linewidth]{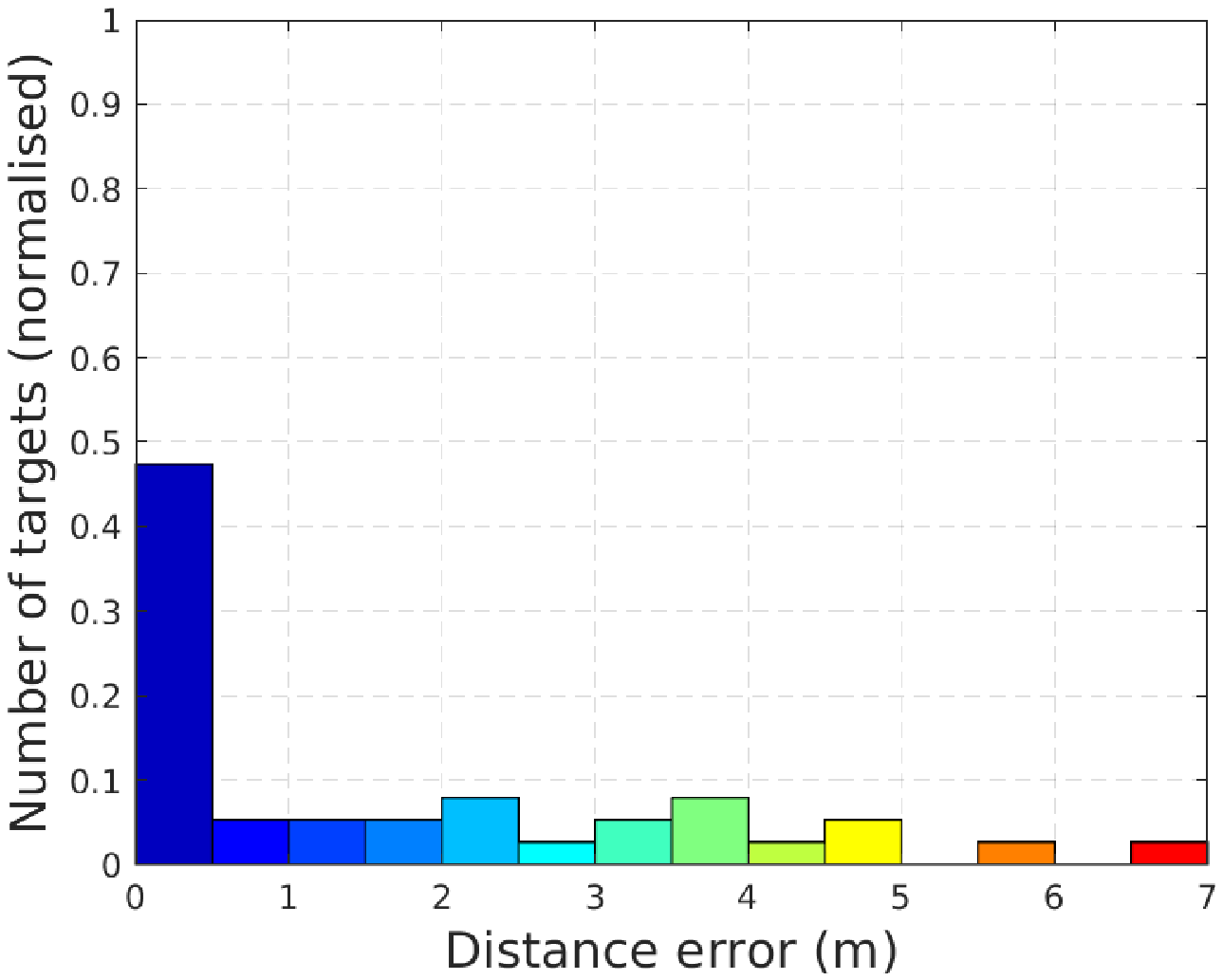}}%
    \qquad
    \subfloat[FAST-LIO2 (QCAT~\spinningpack)]{\includegraphics[width=0.3\linewidth]{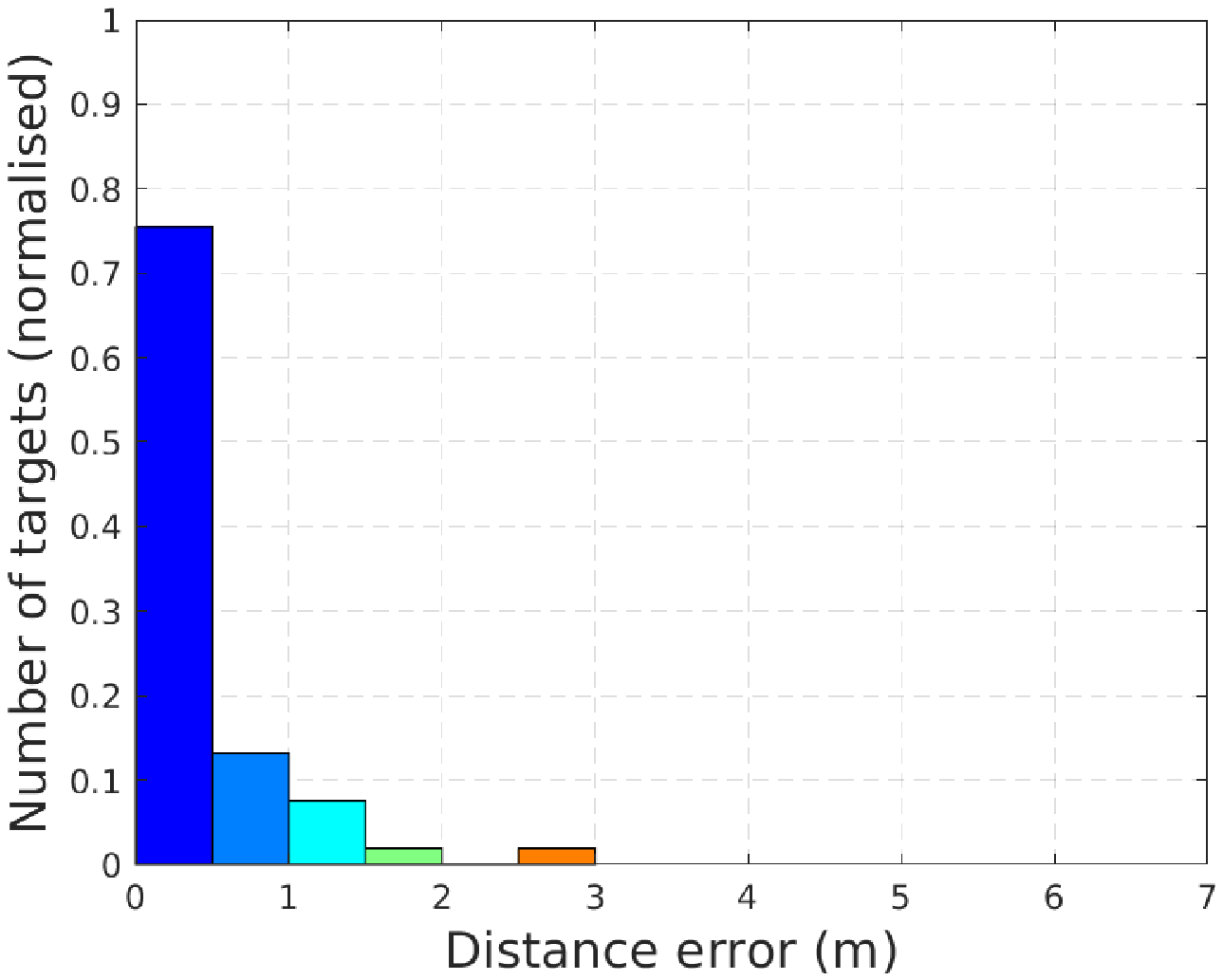}}%

\caption{\small{Error histogram, corresponding to the error statistics reported in~\tabref{tab:results}, over the QCAT \flatpack~dataset (top row) versus QCAT \spinningpack~dataset (bottom row) for Wildcat (left), LIO-SAM (middle) and FAST-LIO2 (right). Number of targets being selected in each method is normalised for better comparison. Note that the results of LIO-SAM and FAST-LIO2 only include a subset of targets before these methods failed.}}
\label{fig:qcat_histograms}
\end{figure*}

\begin{figure}[t]
    \centering
    \subfloat{\includegraphics[width=0.49\linewidth]{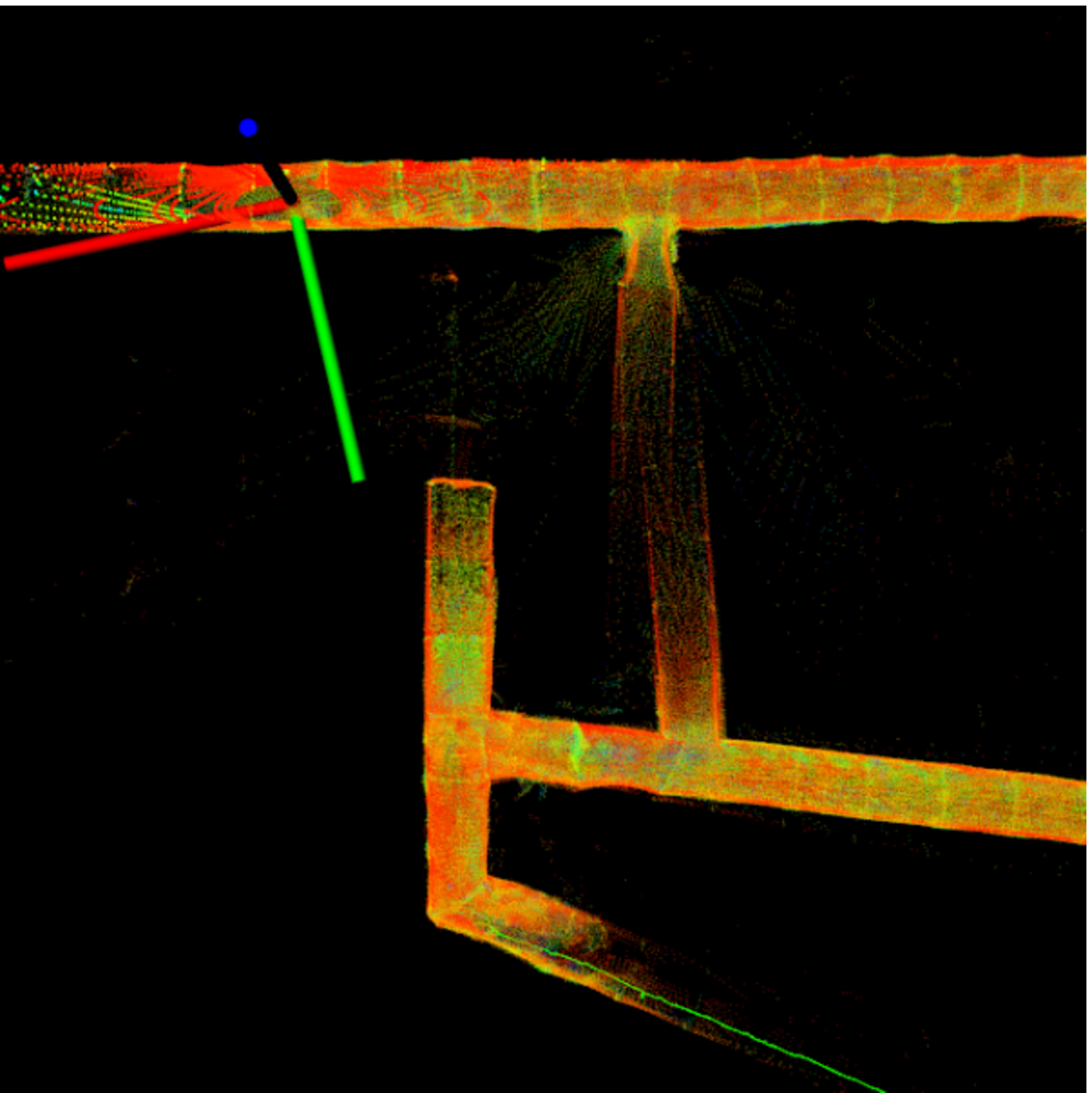}}\
    \subfloat{\includegraphics[width=0.49\linewidth]{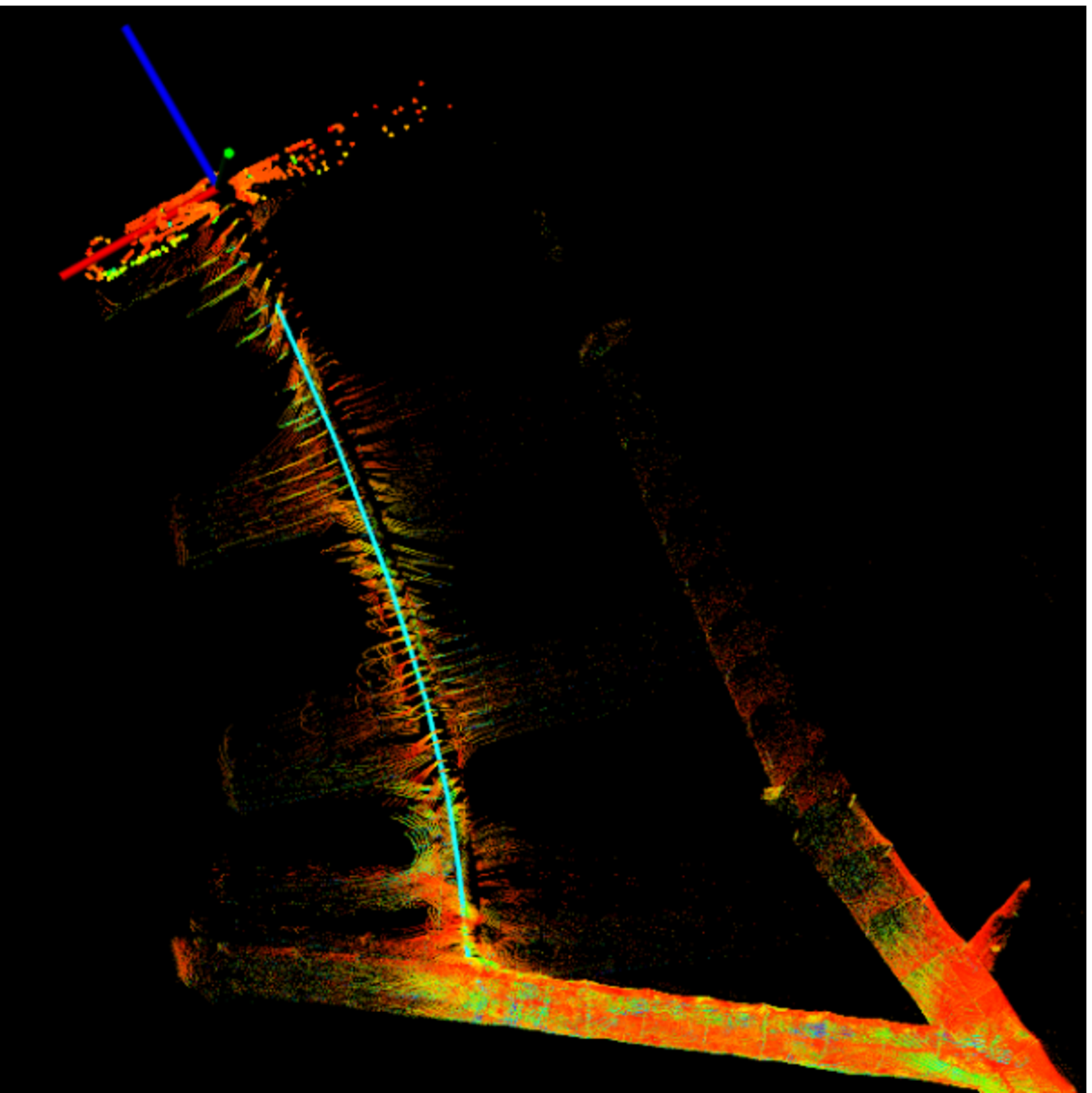}}%
    \caption{\small{Slippage cases for LIO-SAM (left) and FAST-LIO2 (right) through the tunnel on the QCAT~\spinningpack~dataset.}}
\label{fig:tunnel_failures}
\end{figure}

\tabref{tab:results} and \figref{fig:qcat_histograms} show point-to-point distance errors between corresponding pairs of mapped and surveyed (ground truth) targets after an outlier-robust alignment using the M-estimator Sample Consensus (MSAC) method~\cite{torr2000mlesac}.
In addition to the mean and standard deviation (std) of error, \tabref{tab:results} also includes root mean square error (RMSE) computed during alignment. The total number of targets across the QCAT dataset is 63. However, only a subset of these targets could be identified in the maps created by LIO-SAM and FAST-LIO2 due to the fact that both systems exhibited significant error through the tunnel; see
\figref{fig:tunnel_failures}. 
Additionally, LIO-SAM generates very sparse maps compared to Wildcat and FAST-LIO2, hence some of the targets could not be accurately located. Therefore, despite their failures, in \tabref{tab:results} and \figref{fig:qcat_histograms} for LIO-SAM and FAST-LIO2 we report error statistics only for those targets which were mapped accurately enough to be manually identified in the generated 3D maps (see the ``number of targets" column in \tabref{tab:results}).
Unlike LIO-SAM and FAST-LIO2 which could not complete the QCAT experiments, Wildcat performed robustly in all regions and mapped all 63 targets. 

As shown in~\tabref{tab:results}, in the case of the QCAT~\flatpack~dataset, Wildcat's average error is less than half of the average error of LIO-SAM and FAST-LIO2 (without taking into account significant errors in the remaining points after the tunnel). Similarly, on the QCAT~\spinningpack~dataset, Wildcat's average error is about 80\% and 20\% less than that of LIO-SAM and FAST-LIO2, respectively. Furthermore, in both datasets Wildcat has the lowest error standard deviation. These results indicate that, although Wildcat's performance is not dependent on a particular sensor configuration, it can leverage additional information provided by the \spinningpack~compared to \flatpack~to achieve better performance. By contrast, FAST-LIO2 performs poorly in comparison in the~\flatpack~dataset, and LIO-SAM's performance even degrades in the \spinningpack~dataset despite having richer data in comparison to the \flatpack~dataset.
Additionally, \figref{fig:qcat_histograms} shows error histograms normalised by the number of targets identified in each method's map. These histograms show that Wildcat outperforms LIO-SAM and FAST-LIO2 in terms of the fraction of points whose error is below 0.5 m and also maximum error in both the \flatpack{} and \spinningpack{} datasets.

\subsection{Runtime and Memory Analysis}
\label{complexity}
In this part, we report and analyse the runtime and memory consumption of Wildcat in the QCAT~\spinningpack~dataset. The results reported here are collected on a laptop with an Intel Xeon W-10885M CPU. 

\begin{figure}[t]
\centering
\includegraphics[width=1.0\linewidth]{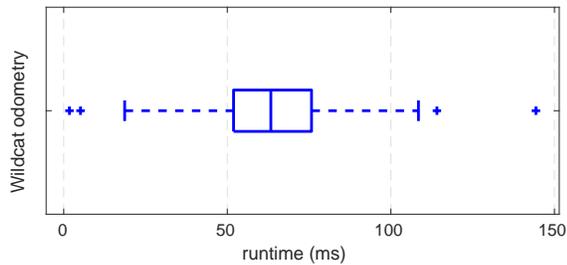}
\caption{\small{Odometry runtime over the QCAT~\spinningpack~dataset.}}
\label{fig:odom_timing}
\end{figure}

\figref{fig:odom_timing} shows the runtime of the main optimisation loop in the odometry module throughout the QCAT~\spinningpack{} dataset. The average runtime is about 63.3 ms (approximately 15 Hz) which shows realtime performance.
On the perception pack's NVIDIA Jetson AGX Xavier onboard computer, the odometry module runs at about 1 to 4 Hz (which is fast enough for processing the current time window before the next one arrives).

\begin{figure}[t]
\centering
\includegraphics[width=1.0\linewidth]{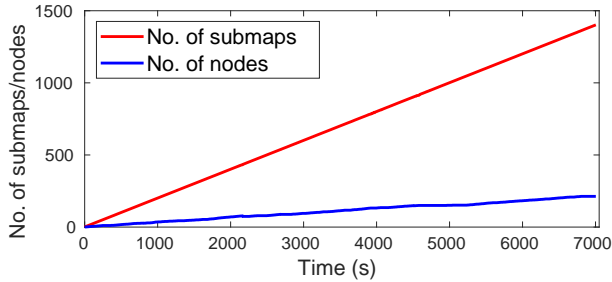}
\caption{\small{Total number of generated submaps versus number of pose graph nodes over the QCAT~\spinningpack~dataset.}}
\label{fig:frames_nodes}
\end{figure}

\begin{figure}[t]
\centering
\includegraphics[width=1.0\linewidth]{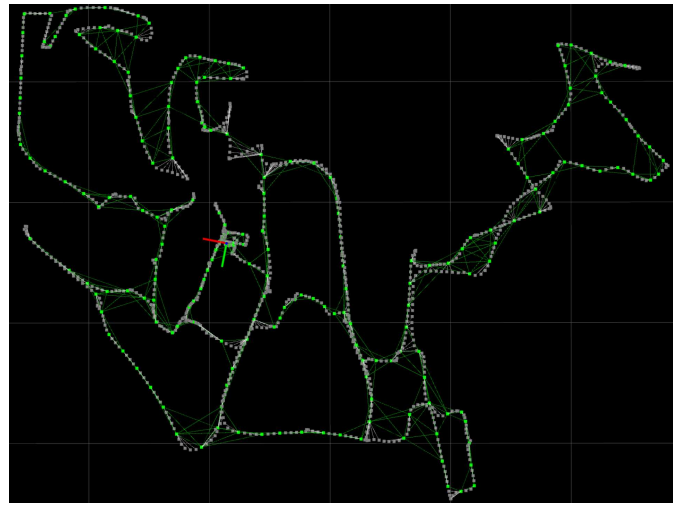}
\caption{\small{Wildcat pose graph generated online over the QCAT~\spinningpack~dataset. PGO effectively merge submaps (grey dots) into nodes (green dots) to reduce the pose graph. Nodes along with the odometry and loop closure edges (green lines) are only used in PGO.}}
\label{fig:pose_graph}
\end{figure}

As we mentioned in Section~\ref{sec:pgo}, one of the key features of our PGO module is the detection and merging of redundant nodes. This enables Wildcat to prevent unnecessary growth of the size of pose-graph optimisation problem over time.
To demonstrate this, we report the total number of submaps generated and the number of nodes in the pose graph over time while running Wildcat online on the~\spinningpack~dataset. As shown in~\figref{fig:frames_nodes}, the total number of submaps is 1402 for the entire experiment, whereas at the end only 213 nodes (obtained by merging submaps) were included in the PGO, resulting in about 85\% reduction of the pose graph size. This key feature thus enables our PGO module to efficiently operate in long-duration missions and large-scale environments. \figref{fig:pose_graph} also shows the distribution of all generated submaps (grey circles) and final set of pose graph nodes (green circles), as well as their corresponding edges.

\begin{figure}[b]
\centering
\includegraphics[width=1.0\linewidth]{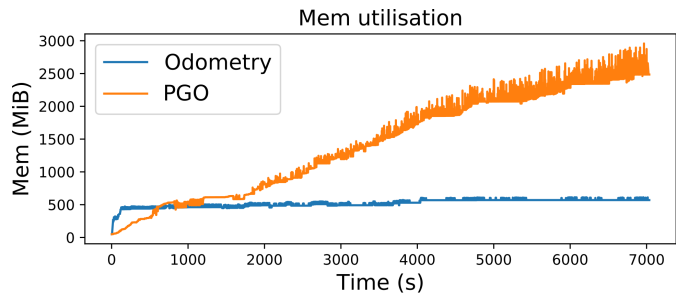}
\caption{\small{Memory usage allocated for the odometry and PGO modules of Wildcat over the QCAT~\spinningpack~dataset.}}
\label{fig:mem_usage}
\end{figure}

Finally, we report the memory consumption of the odometry and PGO modules when running Wildcat online on the QCAT~\spinningpack~dataset. As shown in~\figref{fig:mem_usage}, the odometry memory usage plateaus at the beginning of the run at about 500 MiB. The PGO module consumes more memory than odometry as it needs to store the submaps as the robot explores new areas. However, the total memory consumed by the PGO module for the entire dataset is less than 3 GiB. Additionally, note that as the robot revisits previously explored areas, the memory usage plateaus out due to the fact that PGO merges nodes (and their surfel submaps), thus allowing Wildcat to map large-scale environments more efficiently.

%% file: chapters/conclusion.tex
\section{Conclusion and Future Work}
\label{sec:conclusion}
We presented Wildcat, an online 3D lidar-inertial SLAM system that estimates the robot's 6-DoF motion and efficiently map large-scale environments.
Moreover, we demonstrated its  exceptional robustness and versatility over the state of the art across a wide range of environments and with different types of sensors (VLP-16 in two configurations and OS1-64) and on different platforms (legged, tracked, hand-held, car). Our results indicated that Wildcat outperforms two state-of-the-art methods, especially in challenging environments such as tunnels and corridors. 
The robustness of Wildcat had also been demonstrated at the SubT Challenge Final Event, where Wildcat running in a decentralised fashion on four robots produced the best SLAM results with ``0\% deviation'' according to DARPA.

As for the future work, we plan to improve the resilience and accuracy of the Wildcat odometry module across a wider range of environments and perception systems, incorporate deep learning-based place recognition approaches such as our recent work~\cite{vidanapathirana2021logg3d} into Wildcat for better loop-closure detection, and investigate scaling strategies that can strengthen Wildcat's applicability for multi-agent deployments in larger environments with longer time scales.